\documentclass[pdflatex,sn-mathphys-num]{sn-jnl}

\usepackage{graphicx}%
\usepackage{multirow}%
\usepackage{amsmath,amssymb,amsfonts}%
\usepackage{amsthm}%
\usepackage{mathrsfs}%
\usepackage[title]{appendix}%
\usepackage[table]{xcolor}%
\usepackage{textcomp}%
\usepackage{manyfoot}%
\usepackage{booktabs}%
\usepackage{algorithm}%
\usepackage{algpseudocode}%
\usepackage{listings}%
\usepackage{subcaption}
\usepackage{adjustbox}
\usepackage{algorithmicx}
\usepackage{float}
\usepackage{svg}
\usepackage{tabularx,makecell,array}

\definecolor{background}{HTML}{EEEEEE}
\definecolor{delim}{RGB}{20,105,176}
\colorlet{numb}{magenta}

\lstdefinelanguage{json}{
  basicstyle=\ttfamily\small,
  numbers=left,
  numberstyle=\scriptsize,
  stepnumber=1,
  numbersep=8pt,
  showstringspaces=false,
  breaklines=true,
  frame=single,
  backgroundcolor=\color{background},
  literate=
   *{0}{{{\color{numb}0}}}{1}
    {1}{{{\color{numb}1}}}{1}
    {2}{{{\color{numb}2}}}{1}
    {3}{{{\color{numb}3}}}{1}
    {4}{{{\color{numb}4}}}{1}
    {5}{{{\color{numb}5}}}{1}
    {6}{{{\color{numb}6}}}{1}
    {7}{{{\color{numb}7}}}{1}
    {8}{{{\color{numb}8}}}{1}
    {9}{{{\color{numb}9}}}{1}
    {:}{{{\color{delim}{:}}}}{1}
    {,}{{{\color{delim}{,}}}}{1}
    {\{}{{{\color{delim}{\{}}}}{1}
    {\}}{{{\color{delim}{\}}}}}{1}
    {[}{{{\color{delim}{[}}}}{1}
    {]}{{{\color{delim}{]}}}}{1}
}

\DeclareUnicodeCharacter{202F}{\,}  

\usepackage{cellspace}

\theoremstyle{thmstyleone}%
\theoremstyle{thmstyletwo}%

\theoremstyle{thmstylethree}%

\begin{document}

\title{\textbf{\texttt{HARMON-E} }: \textbf{H}ierarchical \textbf{A}gentic \textbf{R}easoning for \textbf{M}ulti-modal \textbf{O}ncology \textbf{N}otes to \textbf{E}xtract Structured Data}


\author[1]{\fnm{Shashi Kant} \sur{Gupta}}\email{shashi.gupta@triomics.com}

\author[1]{\fnm{Arijeet} \sur{Pramanik}}\email{arijeet.pramanik@triomics.com}
\author[1]{\fnm{Jerrin John} \sur{Thomas}}\email{jerrin.thomas@triomics.com}
\author[1]{\fnm{Regina} \sur{Schwind}}\email{regina@triomics.com}
\author[3]{\fnm{Lauren} \sur{Wiener}}\email{lauren.wiener@mckesson.com}
\author[3]{\fnm{Avi} \sur{Raju}}\email{avi.raju@mckesson.com}
\author[3]{\fnm{Jeremy} \sur{Kornbluth}}\email{jeremy.kornbluth@mckesson.com}

\author[2]{\fnm{Yanshan} \sur{Wang}}\email{yanshan.wang@pitt.edu}
\equalcont{}
\author[3]{\fnm{Zhaohui} \sur{Su}}\email{zhaohui.Su@mckesson.com}
\equalcont{}

\author*[1]{\fnm{Hrituraj} \sur{Singh}}\email{hrituraj@triomics.com}
\equalcont{These authors contributed equally to this work.}

\affil[1]{ \orgname{Triomics}, \city{New York}, \country{USA}}

\affil[2]{ \orgname{University of Pittsburgh}, \city{Pittsburgh}, \country{USA}}
\affil[3]{ \orgname{Ontada}, \city{Boston}, \country{USA}}

\abstract{Unstructured notes within the electronic health record (EHR) contain rich clinical information vital for cancer treatment decision making and research, yet reliably extracting structured oncology data remains challenging due to extensive variability, specialized terminology, and inconsistent document formats. Manual abstraction, although accurate, is prohibitively costly and unscalable. Existing automated approaches typically address narrow scenarios—either using synthetic datasets, restricting focus to document-level extraction, or isolating specific clinical variables (e.g., staging, biomarkers, histology)—and do not adequately handle patient-level synthesis across the large number of clinical documents containing contradictory information. In this study, we propose an \textit{agentic} framework that systematically decomposes complex oncology data extraction into modular, adaptive tasks. Specifically, we use large language models (LLMs) as reasoning agents, equipped with context-sensitive retrieval and iterative synthesis capabilities, to exhaustively and comprehensively extract structured clinical variables from real-world oncology notes. Evaluated on a large-scale dataset of over $400,000$ unstructured clinical notes and scanned PDF reports spanning $2,250$ cancer patients, our method achieves \textbf{an average F1-score of 0.93}, with 100 out of 103 oncology-specific clinical variables exceeding 0.85, and critical variables (e.g., biomarkers and medications) surpassing 0.95. Moreover, integration of the agentic system into a data curation workflow resulted in 0.94 direct manual approval rate, significantly reducing annotation costs.\textbf{ To our knowledge, this constitutes the first exhaustive, end-to-end application of LLM-based agents for structured oncology data extraction at scale.} 
}




\maketitle

\section{Introduction}
\label{sec:introduction}

Traditional efforts toward automating the extraction of structured data from clinical text have relied heavily on rule-based systems or shallow machine learning models (e.g., Conditional Random Fields, Support Vector Machines), each requiring extensive domain-specific feature engineering \cite{uzuner2011, eftimov2017rule, ctakes, deepphe}. In the last decade, however, transformer-based models \cite{vaswani2017}, such as Bidirectional Encoder Representations from Transformers (BERT)\cite{devlin2018bert} and its domain-specific variants like ClinicalBERT \cite{alsentzer2019publicly, lee2020biobert, Preston2023Toward}, have significantly improved upon classical approaches across a range of clinical natural language processing (NLP) tasks. More recently, the approach has shifted from fine-tuning domain specific models \cite{lin2023oncobert, beltagy2019scibert} to using the generalized abilities of frontier large language models (LLMs) like GPT-4 and GPT-5\cite{openai2023gpt4, openai2025gpt5} to extract key concepts from EHR records \cite{Bhattarai2024GPT4}.

Nevertheless, prior works often assume relatively uniform data (such as a single cancer diagnosis or one type of note), single-document inputs, or a very limited set of concepts. Published studies have demonstrated entity extraction for discrete variables such as tumor stage from pathology reports \cite{abedian2021automated, kefeli2024tnm}, receptor or genomic biomarkers from pathology or genomic reports \cite{gauthier2022rwe, pironet2021receptor}, and initial regimens or lines of therapy for select cancers \cite{zeng2021initialtx, meng2021lot}. However, these approaches extract variables independently without synthesizing findings across multiple documents to resolve contradictions or complete partial information. This study is distinct in that it targets patient-level synthesis across heterogeneous, longitudinal records, where oncology-specific concepts must be inferred by collating and contextualizing information scattered across multiple notes and data types, rather than relying on single-step extraction from a uniform document.

To address these challenges, we introduce \textbf{\texttt{HARMON-E} }:
\textbf{H}ierarchical \textbf{A}gentic \textbf{R}easoning for \textbf{M}ulti-source \textbf{O}ncology \textbf{N}otes to \textbf{E}xtract structured data

\begin{figure}
    \centering
    \includegraphics[width=1\linewidth]{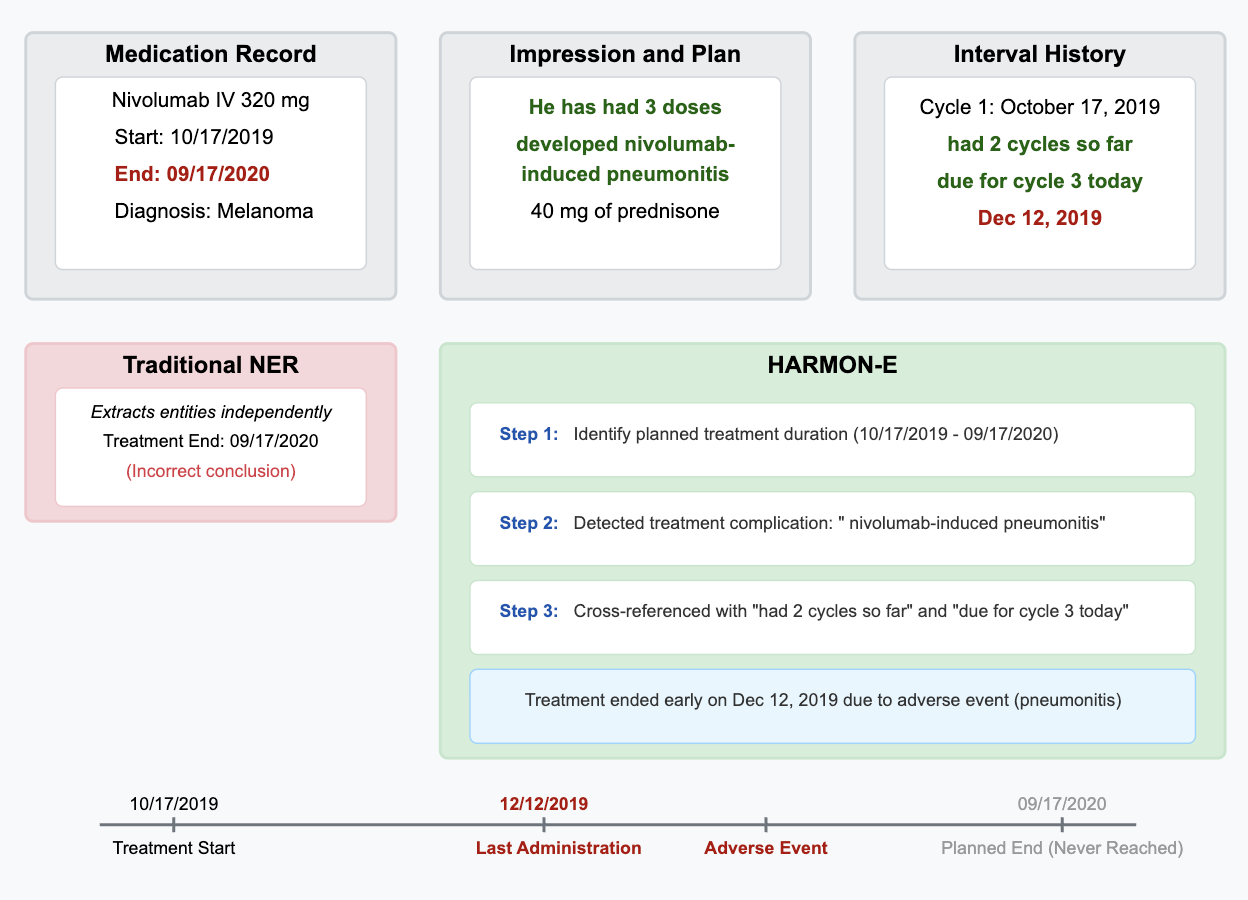}
    \caption{\textbf{Comparative analysis of \texttt{HARMON-E} versus traditional Named Entity Recognition (NER) approaches for medical document processing}. A traditional NER system (lower left, red box) would incorrectly extract the planned end date (09/17/2020) without contextual understanding. In contrast, our agentic system (lower right, green box) performs multi-step reasoning: first identifying the planned treatment duration, then detecting the adverse event (pneumonitis), cross-referencing information about completed treatment cycles, and finally concluding that treatment actually ended on December 12, 2019 due to the adverse event—a conclusion impossible with traditional single-pass methods. (This is not real patient data and is for illustrative purposes only.) }
\label{fig:example}
\end{figure}

\begin{figure}
    \centering
    \includegraphics[width=1\linewidth]{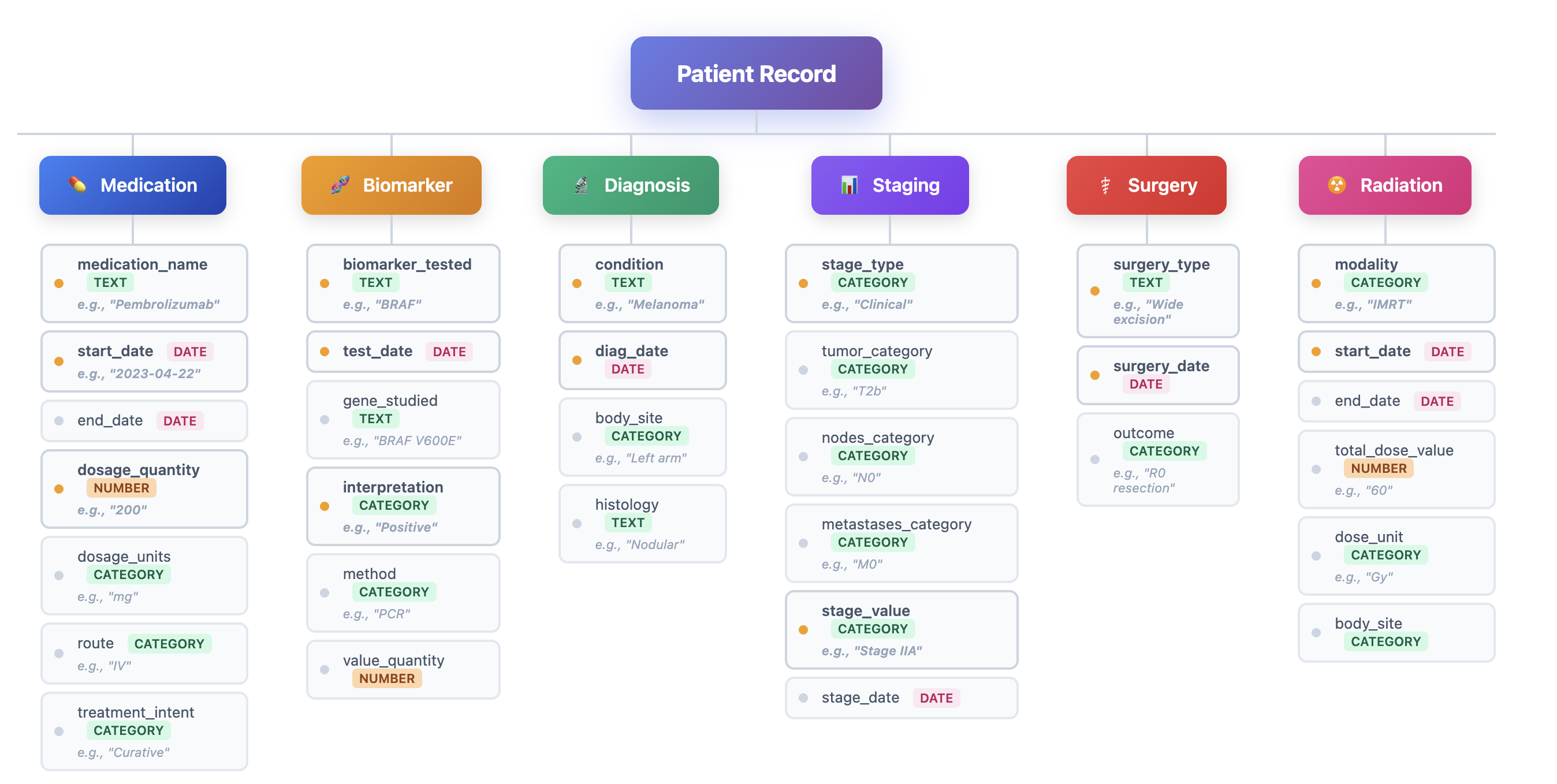}
    \caption{ The tree structure illustrates the decomposition of a patient record into six primary entity categories (Medication, Biomarker, Diagnosis, Staging, Surgery, and Radiation), each containing multiple typed attributes. Representative examples from real oncology data are provided in italics below each attribute. This structured schema ensures consistency across heterogeneous clinical documentation and enables validation of extracted data against clinical standards. The complete model encompasses 16 entity types with 103 distinct attributes (subset shown for clarity).}
    \label{fig:entity_schema}
\end{figure}

Our framework systematically decomposes oncology data extraction into modular, iterative steps that leverage large language models (LLMs) as ``reasoning agents'' that interleave retrieval with multi-step inference to reconcile conflicting evidence, normalize temporal references, and derive implicit variables not explicitly stated in the text - such as inferring treatment discontinuation dates from adverse event narratives (Figure \ref{fig:example}). Concretely, \texttt{HARMON-E}  combines context-aware indexing, adaptive retrieval methods (including vector-based search and rule-based actions), and LLM-driven data synthesis into a unified workflow. By adopting a hierarchical, agentic approach, we mitigate common pitfalls in extracting complex, interdependent oncology data and achieve end-to-end alignment with standardized pre-defined set of clinical variables comprising 103 attributes across 16 entity types-including Biomarker, Medication, Diagnosis, Staging, Surgery and Radiation entities as outlined in Table \ref{tab:merged_entities} and Figure {\ref{fig:entity_schema}}.

We validate \texttt{HARMON-E}  on a large-scale real-world dataset of over 400,000 unstructured clinical notes and scanned PDFs belonging to 2,250 cancer patients. Our evaluation covers 103 oncology-specific variables—ranging from histopathological findings and biomarker statuses to treatment patterns and disease progression. \texttt{HARMON-E}  consistently delivers high accuracy, with 100 out of 103 variables exceeding an F1-score of 80\%, and crucial fields such as biomarker assessments and medication data surpassing 95\% accuracy. Beyond these metrics, when integrated into a data curation platform, our system demonstrates a direct manual approval rate of 94.1\%, substantially reducing human annotation burden without sacrificing data quality.

Our contributions can be summarized as follows -

\begin{enumerate}
    \item \textbf{High-accuracy LLM-based agentic workflows.} We provide the first evidence that large language models, organized into an agentic workflow, can exceed 95\% F1-score on key oncology concepts at scale, validated against a corpus of 940{,}923 data points derived from 400{,}000 clinical notes covering 2{,}250 patients.

    \item \textbf{Real-world integration and user acceptance.} We also demonstrate that when integrated into an interactive data curation platform, 94\% of the extracted data points get directly accepted by professional oncology abstractors without modification, substantially reducing manual review time.
    
    \item \textbf{Novel evaluation framework.} We propose an evaluation methodology which moves beyond traditional named-entity recognition metrics by aligning performance assessment with the real-world requirements of oncology data curation and quality monitoring.

\end{enumerate}

\section{Related Work}
\label{sec:related_work}

\textbf{Early Clinical NLP and Rule-Based Methods:} Early clinical NLP relied primarily on rule-based systems, utilizing domain-specific lexicons and hand-crafted patterns \cite{friedman2004, chapman2001, ctakes}. These systems, though precise in certain scenarios, required substantial manual effort and were brittle when encountering variability \cite{uzuner2012evaluating, uzuner2011}. Notable examples include MedXN for medication extraction \cite{sohn2014medxn}, comprehensive clinical NLP toolkits like CLAMP \cite{soysal2018clamp}, fracture identification from radiology reports \cite{tibbo2019radiology}, and sudden cardiac death risk factor extraction \cite{moon2019extraction}. Statistical machine learning models, notably Conditional Random Fields (CRFs) and Support Vector Machines (SVMs), subsequently improved performance on clinical text tasks like de-identification and entity extraction \cite{uzuner2011}. Early language modeling approaches showed promise for identifying relevant information in clinical notes \cite{zhang2014automated}, though still required task-specific adaptation. Nevertheless, adapting these models to heterogeneous oncology datasets posed significant challenges due to diverse terminologies and complex documentation \cite{murff2011automated}. The evolution of clinical information extraction methods has been comprehensively reviewed by Wang et al. \cite{wang2018clinical}, documenting the systematic transition from rule-based to neural approaches.

\textbf{Deep Learning and Transformer Models in Clinical NLP:} Transformer architectures, particularly BERT \cite{vaswani2017, devlin2018bert}, significantly advanced clinical NLP by providing powerful language representations. Domain-specific models such as BioBERT \cite{lee2020biobert}, ClinicalBERT \cite{alsentzer2019publicly}, and SciBERT \cite{beltagy2019scibert} further enhanced clinical NLP tasks, achieving state-of-the-art performance in clinical entity recognition and relation extraction \cite{peng2019transfer, si2019deep}. Comprehensive evaluations of these models on biomedical text-mining tasks demonstrated their superiority over traditional approaches \cite{peng2019empirical}. However, oncology data, characterized by extensive and contextually complex documentation, poses unique challenges inadequately addressed by traditional single-pass transformers. Approaches employing hierarchical transformers and multi-document summarization have attempted to mitigate these limitations \cite{ding2020cogran, fabbri2019multi, Preston2023Toward, Adamson2023Approach} but still fall short of fully replicating human abstraction.

\textbf{Large Language Models in Healthcare:} Recent advances in large language models (LLMs), including GPT-4, and GPT-5\cite{openai2023gpt4, openai2025gpt5}, have demonstrated impressive capabilities in summarization, medical question answering, and clinical decision support \cite{kung2023performance, agrawal2023large}. This has led to several research works exploring the capabilities of these models in extracting structured data from notes \cite{Bhattarai2024GPT4, wong2025universal, porter2024llmd}. Recent multi-institutional efforts have shown promise in extracting social determinants of health from clinical notes using LLMs, achieving F1 scores over 0.9 \cite{keloth2025social}. Work on extracting functional status information, including mobility assessments, from clinical notes has demonstrated the potential of LLMs for capturing complex clinical concepts \cite{fu2024federated, kaster2025automated}. Despite this progress, single-pass LLM approaches struggle with multi-document EHR data, potentially losing critical context or misinterpreting conflicting information. Moreover, most research work so far has been limited to a limited set of oncology concepts \cite{Adamson2023Approach, Preston2023Toward, wong2025universal, stuhlmiller2025scalable} instead of scaling it to a comprehensive RWD dataset dictionary.

\textbf{Agentic and Iterative NLP Frameworks:} Recent NLP frameworks have increasingly adopted iterative reasoning approaches. The ReAct framework introduced by Yao et al. \cite{yao2022react} enables LLMs to interleave reasoning and retrieval actions, improving accuracy on complex reasoning tasks. Techniques such as chain-of-thought prompting \cite{wei2022chain} and self-consistency decoding \cite{wang2022self} further enhance multi-step reasoning capabilities. Additionally, integrating external tools or retrieval systems, as explored in Toolformer \cite{schick2023toolformer} and web-augmented retrieval methods \cite{nakano2021webgpt}, has shown promise. Yet, systematically integrating these approaches for clinical NLP, particularly oncology-specific data extraction, remains an open challenge due to accuracy and validation constraints.

\textbf{Autonomous \& Agentic LLM Systems in Oncology:}
Oncology specific AI research has also begun moving from single-prompt LLM use toward \emph{agentic} pipelines that interleave reasoning and tool calls.
Ferber~\emph{et al.} developed an autonomous GPT-4–driven clinical-decision agent that orchestrates vision models, knowledge bases, and web search to solve multimodal oncology cases, raising accuracy from 30\% (plain GPT-4) to over 87\% on complex vignettes~\cite{ferber2025autonomous}. Sandhu~\emph{et al.} proposed an open-source modular framework that couples rule-based components with an agentic layer to generate comprehensive breast-cancer notes and benchmark treatment recommendations against NCCN guidelines~\cite{sandhu2025agentic}. Outside healthcare, Zhang and Elhamod proposed \textit{Data-to-Dashboard}, a multi-agent architecture that detects domain context, extracts concepts, performs iterative self-reflection, and automatically builds analytic dashboards~\cite{zhang2025data_dashboard}.
While these works demonstrate the promise of hierarchical agents, they tackle relatively \emph{narrow} tasks (tens of test cases, disease-specific notes, or generic analytics) rather than end-to-end patient-level abstraction across thousands of heterogeneous documents.

\textbf{Domain-Specific LLMs for Oncology.}
Parallel efforts tailor foundation models and pipelines to oncology data.
\emph{CancerBERT} pretrains BERT on cancer corpora with a cancer-specific vocabulary and reports significant gains for extracting breast-cancer phenotypes (e.g., receptor status, site/laterality) over general and clinical BERT baselines from EHR notes and pathology reports\cite{zhou2022cancerbert}.
\emph{OncoBERT} explores transfer learning on oncology notes for outcome prediction and phenotype structuring, reporting improvements on site and clinical T-staging tasks and highlighting interpretability considerations for radiation oncology workflows \cite{lin2023oncobert,Preston2023Toward}.
Beyond encoders, registry-oriented systems deliver patient-level abstraction: the original \emph{DeepPhe} extracts phenotypes across entire EMRs, while \emph{DeepPhe-CR} exposes API services integrated into cancer-registrar tools for
computer-assisted abstraction, with usability studies and deployment guidance for
registry workflows \cite{deepphe,hochheiser2023deepphecr}.
Patient-level staging at treatment initiation has also been demonstrated in Veterans Affairs data using rule-based NLP with validated roll-up for multiple myeloma \cite{vaMMNLP2024}. In parallel, registry–EHR fusion cohorts constructed via NLP underscore the need for multi-source abstraction \cite{ling2019mbc}.
A complementary line targets narrow modalities or procedures—e.g., \emph{Woollie}, a radiology-focused LLM trained on 39k impressions that attains strong progression-prediction AUROC of $0.97$ for progression prediction and outperforms general LLMs on medical benchmarks, and hybrid agentic pipelines coupling rules with GPT-4-Turbo for detailed spine-surgery variable extraction \cite{zhu2025woollie,dagli2024ai_chart_review}.
Orthogonal oncology IE efforts focus on single-entity extraction from pathology and clinical notes, including TNM staging and receptor status \cite{abedian2021automated,kefeli2024tnm,pironet2021receptor}, systemic treatment identification and line-of-therapy inference \cite{zeng2021initialtx,meng2021lot}, and broader RWD curation reviews and tooling \cite{gauthier2022rwe,Zeng2019NLPPhenotyping}.

Most oncology LLM studies target narrow concept sets rather than comprehensive data dictionaries \cite{Adamson2023Approach,Preston2023Toward,wong2025universal,stuhlmiller2025scalable}. These systems typically optimize for task- or modality-specific accuracy using single-pass encoders or rule-based components; they provide limited mechanisms for cross-entity dependency resolution, large-scale deduplication across documents, and curator-ready validation—gaps our agentic workflow aims to close.
 
To overcome this, \texttt{HARMON-E} aim to mirror the iterative reasoning process of human abstractors. By combining state-of-the-art linguistic models and structured retrieval actions, our proposed approach promises robust, scalable extraction from heterogeneous oncology records. To our knowledge, this is the first agentic LLM framework that delivers both \emph{breadth} (comprehensive data dictionary) and \emph{depth} (patient-level consistency) at scale, bridging the gap between prototype agents or niche LLMs and production-grade oncology data curation.

\begin{table}[htbp]
\centering
\scriptsize
\caption{Clinical Entities, Definitions, and Corresponding Attributes}

\rowcolors{2}{gray!10}{white}
\begin{tabular}{p{2.8cm} p{5cm} p{7.3cm}}
\toprule
\rowcolor{blue!20}
\textbf{Entity Name} & \textbf{Definition of Entity} & \textbf{Attributes} \\
\midrule

\textbf{Biomarker} 
& Consolidated entity for all biomarker testing, including genetic biomarkers, microsatellite instability, copy number alterations, rearrangements, tumor marker tests, and tumor mutation burden.
& \texttt{biomarker\_test\_date}, \texttt{result\_date}, 
  \texttt{biomarker\_tested}, \texttt{gene\_studied}, \texttt{gene1}, 
  \texttt{copy\_number\_type}, \texttt{method}, \texttt{code}, 
  \texttt{aminoacid\_change}, \texttt{aminoacid\_changetype}, 
  \texttt{stain\_percent}, \texttt{value}, \texttt{value\_quantity}, 
  \texttt{value\_unit}, \texttt{interpretation}, 
  \texttt{molecular\_abnormal} \\

\textbf{Biopsy} 
& Describes the biopsy specimen (e.g., liquid or tissue) and dates. 
& \texttt{type}, \texttt{result\_date}, \texttt{ordered\_date}, \texttt{collect\_date}, \texttt{insufficient\_tissue} \\

\textbf{Clinical Trial} 
& Derived from clinical trial attributes. Indicates patient enrollment in a trial. 
& \texttt{ct\_flag}, \texttt{ct\_start\_date}, \texttt{ct\_end\_date} \\

\textbf{Comorbidities} 
& Captures health conditions (Charlson Comorbidity Index) co-existing with primary cancer.
& \texttt{comorb\_date}, \texttt{comorb\_condition}, \texttt{comorb\_condition\_present} \\

\textbf{Distant Metastasis}
& Indicates metastatic spread beyond the primary site after initial diagnosis.
& \texttt{status\_date}, \texttt{status}, \texttt{associated\_diagnosis}, \texttt{body\_site} \\

\textbf{Diagnosis}
& Details about the primary diagnosis, including histology and site.
& \texttt{diag\_date}, \texttt{condition}, \texttt{body\_site}, \texttt{histology} \\

\textbf{Family History}
& Details the family cancer history for a patient.
& \texttt{relationship}, \texttt{condition}, \texttt{onset} \\

\textbf{Imaging}
&  Describes imaging services a patient has received.
& \texttt{start\_date}, \texttt{modality}, \texttt{body\_site} \\

\textbf{Medication}
&  Captures systemic treatments (drugs, therapies), start/end dates, routes, etc.
& \texttt{status}, \texttt{start\_date}, \texttt{end\_date}, \texttt{treatment\_intent}, 
  \texttt{medication}, \texttt{route}, \texttt{termination\_reason},
  \texttt{baseline\_dosage}, \texttt{baseline\_dosage\_units}, 
  \texttt{baseline\_dosage\_quantity}, \texttt{baseline\_dosage\_freq},
  \texttt{baseline\_dosage\_duration}, \texttt{baseline\_cycle\_length},
  \texttt{dose\_change\_reason}, \texttt{changed\_dosage},
  \texttt{changed\_dosage\_units}, \texttt{changed\_dosage\_quantity},
  \texttt{changed\_dosage\_freq}, \texttt{changed\_dosage\_duration},
  \texttt{changed\_cycle\_length}, \texttt{dose\_formula},
  \texttt{dose\_formula\_unit}, \texttt{treatment\_sequence} \\

\textbf{Nicotine Use Status} 
& Describes patient’s smoking or nicotine usage.
& \texttt{code}, \texttt{type}, \texttt{use}, \texttt{use\_unit}, \texttt{use\_frequency}, \texttt{start\_date}, \texttt{end\_date} \\

\textbf{Patient Status}
& Covers demographics, vital status, and disposition.
& \texttt{vital\_status}, \texttt{last\_contact\_date}, \texttt{hospice\_date}, \texttt{relapse\_flag} \\

\textbf{Radiation}
& Captures radiation therapy details.
& \texttt{modality}, \texttt{start\_date}, \texttt{end\_date}, \texttt{total\_dose\_delivered\_value}, \texttt{total\_dose\_delivered\_unit}, \texttt{fractions\_delivered}, \texttt{body\_site} \\

\textbf{Recurrence Status}
& Tracks local or metastatic recurrence of disease.
& \texttt{status\_date}, \texttt{status}, \texttt{associated\_diagnosis}, \texttt{body\_site} \\

\textbf{Staging}
& Describes TNM staging at diagnosis.
& \texttt{stage\_date}, \texttt{stage\_type}, \texttt{tumor\_category}, \texttt{nodes\_category}, \texttt{metastases\_category}, \texttt{stage\_value} \\

\textbf{Surgery}
& Dates and types of surgical procedures received.
& \texttt{surgery\_date}, \texttt{surgery\_type}, \texttt{outcome} \\

\textbf{Surgery Observations}
&  Specific to surgical procedures and outcomes.
& \texttt{observe\_date}, \texttt{code}, \texttt{value}, \texttt{value\_units} \\

\bottomrule
\end{tabular}
\label{tab:merged_entities}
\end{table}

\section{Methods}

\subsection{Problem Formulation}
\begin{figure}
    \centering
    \includegraphics[width=1\linewidth]{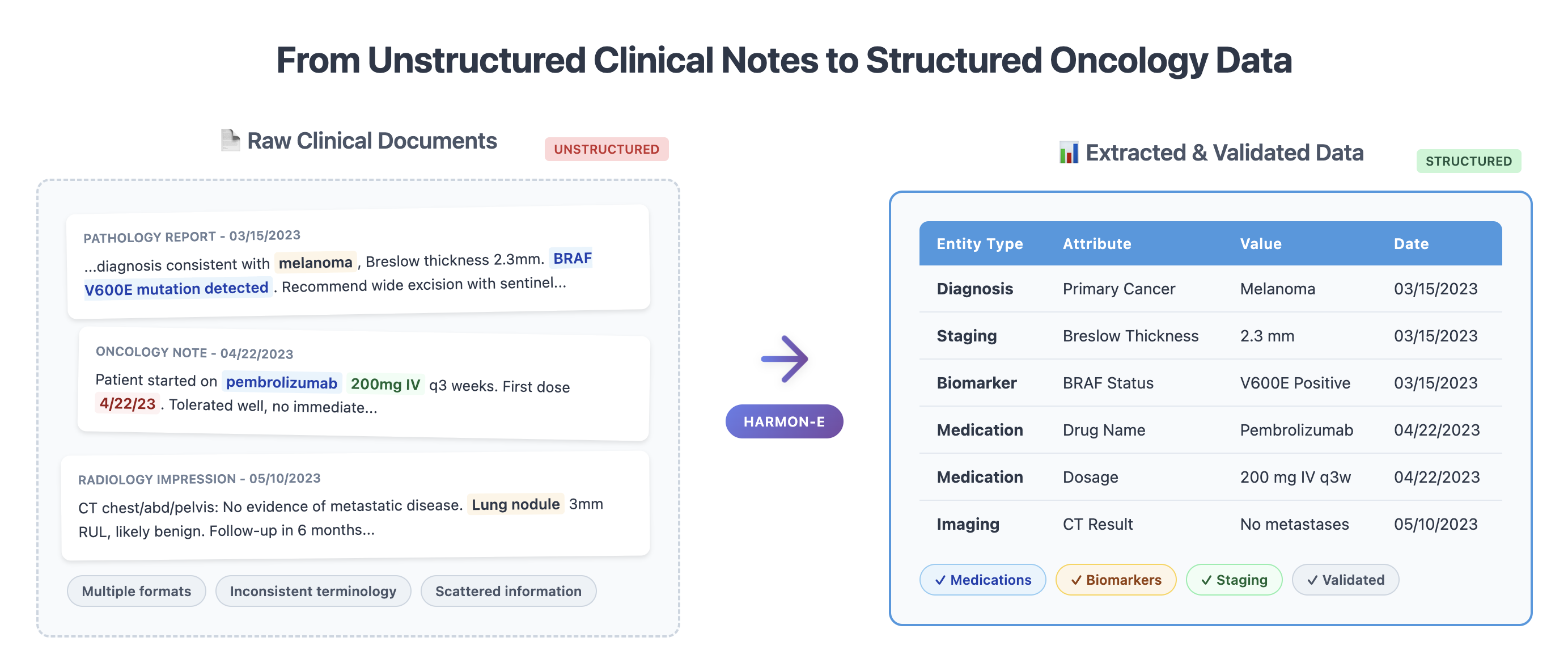}
\caption{\textbf{The HARMON-E transformation pipeline.} Unstructured clinical documents from multiple sources (left) containing medications, biomarkers, staging information, and other oncology data are processed through the HARMON-E agentic extraction pipeline (center) to produce standardized, structured database entries (right). Each piece of clinically relevant information is extracted, validated, and organized into predefined entity-attribute pairs suitable for clinical research and decision support. The system processes heterogeneous inputs including progress notes, pathology reports, radiology impressions, and scanned PDFs, transforming approximately 180 documents per patient into comprehensive structured records.}
\label{fig:problem_formulation}
\end{figure}

We consider a set of patients 
\[
\mathcal{P} = \{p_1, p_2, \ldots, p_n\}
\]
and a predefined collection of \emph{clinical entities} relevant to oncology (e.g., \texttt{Biomarker}, \texttt{Cancer Related Medication}, \texttt{TNM Stage Group}). Each entity $E_j$ consists of a collection of \emph{attributes}
\[
\mathcal{A}(E_j) \;=\; \{\, a_{j,1},\, a_{j,2},\,\ldots,\, a_{j,k_j} \}, \qquad k_j:=|\mathcal{A}(E_j)|.
\]
where each attribute can be a categorical label (with a fixed value set), a date, a numeric value, or unstructured text. Here, \(k_j\) denotes the number of attributes defined for entity \(E_j\) and can differ by entity (see Table~\ref{tab:merged_entities}). Given a corpus of unstructured or semi-structured oncology documents 
\[
\mathcal{D}_i = \{\,d_1, d_2, \ldots, d_m\}\quad \text{for each patient }p_i,
\]
our goal is to automatically \emph{extract} all valid instances of these entities. Equivalently, we want a structured output $\mathcal{S}_i$ for every patient $p_i$ containing all detected entities $\{\,E_1, \ldots, E_J\}$ and their attribute values. Thus,
\[
\mathcal{S}_i \;=\; \bigcup_{j=1}^{J} \hat{E}_j,
\]
where $J$ is the total number of clinical entity types considered, and each $\hat{E}_j$ is a set of extracted instances of type $E_j$. Table~\ref{tab:merged_entities} lists 16 oncology entity types considered in this study.

We evaluate the quality of extraction by comparing the system outputs against a gold standard dataset. For each entity type, we assess:
\begin{itemize}
    \item \textbf{Entity-Level Recall:} Do we capture \emph{all} correct entity instances mentioned in patients' notes?
    \item \textbf{Entity-Level Precision:} Do we avoid producing extra or incorrect entity instances?
    \item \textbf{Attribute-Level Accuracy:} Conditioned on having a correct entity instance, are all attributes (e.g., dates, biomarker results) accurate?
\end{itemize}
We further perform alignment between predicted entity instances and ground-truth references (e.g., via root-based or weighted matching) to compute these metrics systematically as discussed in the Section \ref{sec:eval}.

\subsection{Document ingestion and normalization}
\label{subsec:ingestion}
Scanned PDFs are transcribed to page-level Markdown using a vision–language model fine-tuned on clinical documents. We then invoke two LLM calls for (i) page-to-document segmentation and doc typing, and (ii) metadata extraction (encounter date, report title, identifiers). The resulting normalized text feeds the retrieval–synthesis–collation pipeline.

\subsection{Architecture}
\label{sec:architecture}
\begin{figure}
    \centering
    \includegraphics[width=1\linewidth]{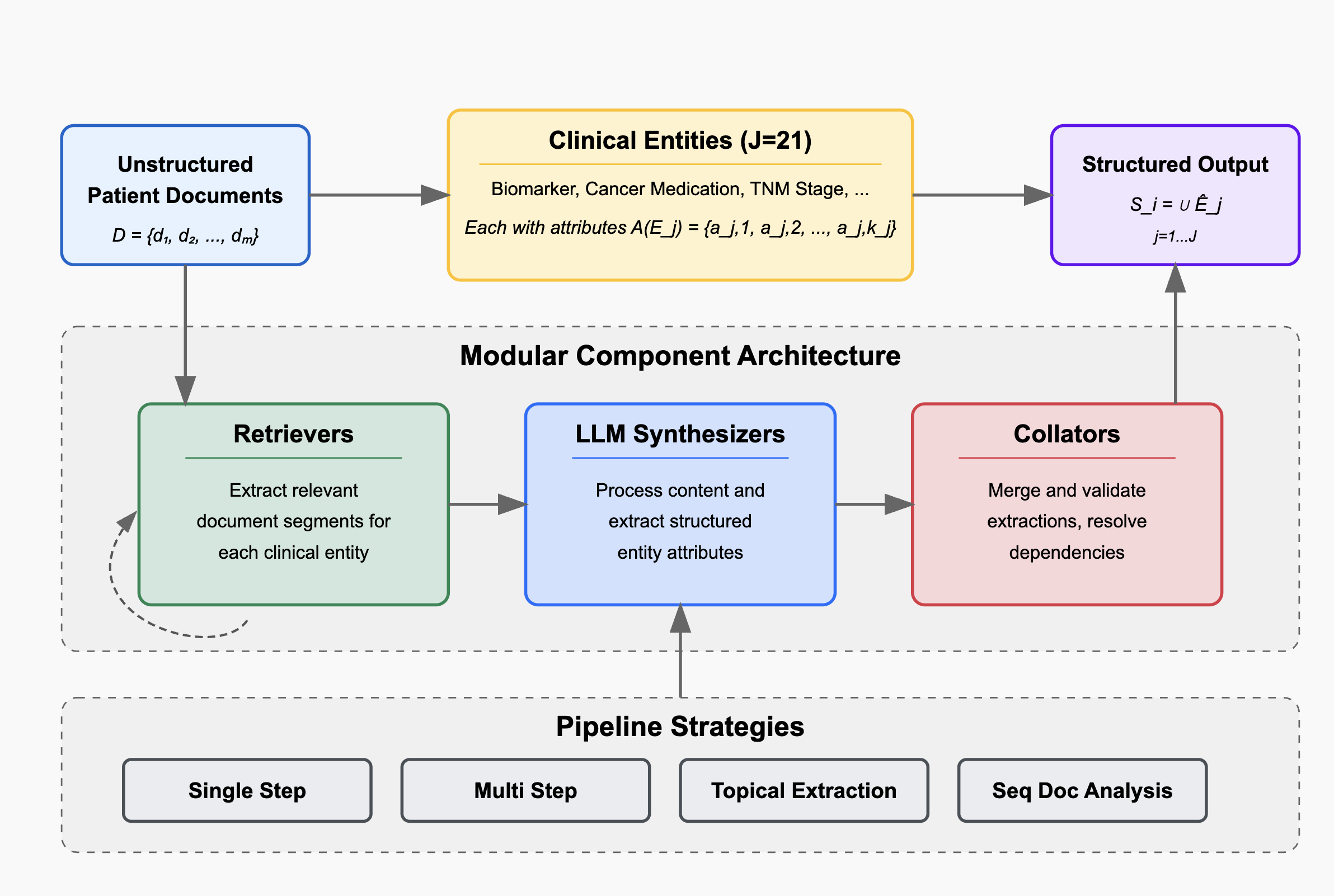}
    \caption{
        \textbf{System workflow of \texttt{HARMON-E}.}
        The system accepts raw, unstructured patient documents (HTML notes and/or scanned PDFs normalized as in Sec.~\ref{subsec:ingestion}) and processes them through three main components: \emph{Retrievers}, \emph{LLM Synthesizers}, and \emph{Collators}. 
        Retrievers extract relevant segments for each targeted oncology entity; 
        LLM Synthesizers transform these segments into structured attribute-value pairs; 
        and Collators merge, validate, and resolve any dependencies among the extracted entities.
        The framework supports multiple pipeline strategies---from single-step to multi-step or topical extraction---accommodating a diverse range of real-world workflows.
        The final output is a patient-level structured data record spanning multiple oncology concepts (e.g., biomarkers, medications, TNM staging).
    }
    \label{fig:architecture}
\end{figure}
We design a modular, domain-focused architecture for extracting oncology-specific data variables at scale. The architecture decomposes the problem into three \emph{components} and integrates them into one of several \emph{pipeline} strategies, each specializing in different document-processing scenarios.

\subsubsection{Key Features}
\begin{itemize}
    \item \textbf{Modular Component Architecture:} We expose interchangeable building blocks (\textit{retrievers}, \textit{LLM synthesizers}, and \textit{collators}) to accommodate different data formats and use cases.
    \item \textbf{Multiple Pipeline Strategies:} We implement multiple component combinations (\emph{Single-Step}, \emph{Multi-Step}, \emph{Topical Extraction}, \emph{Sequential Document Analysis}), each with distinct retrieval-extraction flows.
    \item \textbf{Flexible Configuration System:} Users can seamlessly modify pipeline parameters (e.g., queries, prompts, pattern matchers) without altering core logic.
    \item \textbf{Strong Typing \& Validation:} Entity attributes are rigorously typed (categorical, date, numeric, etc.) to ensure consistent outputs and validate data integrity.
    \item \textbf{Dependency Management for Complex Extractions:} Certain extractions rely on previously resolved attributes (e.g., a medication collator may need a confirmed diagnosis date). Our architecture manages these dependencies explicitly.
\end{itemize}

\subsubsection{Core Data Models}
At the heart of the system are \emph{entities} and \emph{attributes}. Each entity $E_j$ includes:
\[
\mathcal{A}(E_j) \;=\; \{\, (n_{j,i},\, t_{j,i},\, V_{j,i}) \,\}_{i=1}^{k_j},
\]
where $n_{j,i}$ is an attribute name (e.g., \texttt{biomarker\_tested}), $t_{j,i}$ is its type (e.g., \texttt{Date}, \texttt{Integer}, or \texttt{Categorical}), and $V_{j,i}$ is an optional finite set of valid values for categorical attributes (e.g., \{\texttt{Positive, Negative}\}).

We also incorporate \emph{strong typing} by rejecting any extractions that violate the declared schema (e.g., a numeric field with an invalid string). In practice, this ensures consistent representation across different pipeline stages.

\subsubsection{Main Components}

\paragraph{1.~Retrievers}
A retriever receives text input and yields a list of \emph{candidate chunks} (snippets) relevant to the extraction goal. Clinical notes are chunked deterministically into sentence-bounded windows, each capped at $M$ characters, with a one-sentence overlap between consecutive windows. We consider two broad categories:

\begin{itemize}
    \item \textbf{Vector Retriever}: Embedding-based similarity search to identify chunks that semantically match a query (e.g., a disease name or biomarker). 
    \item \textbf{Regex Retriever}: Regex-based pattern matching to capture well-defined textual cues (e.g., specific drug names, standard biomarkers). 
\end{itemize}

\noindent\textbf{How queries are defined.}
Retrieval queries are \emph{entity-conditioned}. For a target entity $E_j$ and attribute subset $\mathcal{S}$, the user configures $\{q_1,\dots,q_m\}$, where $m$ is the number of queries generated for that extraction call.

\noindent\textbf{Example queries.}
\emph{Diagnosis (surgery):} “Has the patient undergone any resection?”, “Is there a mention of mastectomy?”, “What is the surgery date?” \\
\emph{Biomarker (BRAF):} “When was BRAF last tested?”, “What was the BRAF test result?”, “How did the lab interpret the BRAF result?”

Algorithmically, a \emph{Vector Retriever} can be summarized as:
\begin{algorithm}[H]
\caption{Vector Retriever (Abstract)}
\begin{algorithmic}
\Require A set of chunk texts, an embedding function $\phi(\cdot)$, a user query $q$, and retrieval size $k$.
\State Compute the embedding $\phi(q)$ for the query.
\State For each chunk text $d$, compute $\phi(d)$.
\State Rank chunks by cosine similarity $\langle \phi(q), \phi(d) \rangle$.
\State Return top $k$ most similar chunks.
\end{algorithmic}
\end{algorithm}

For each query $q_r$, \texttt{top\_k} ($k_r$) is the number of highest-scoring chunks that is retained after ranking; larger $k_r$ increases recall at the cost of more noise and latency.

\textit{Examples:} surgery-identification queries use $k_r{=}16$ to catch sparse mentions; surgery-detail queries (e.g., date, margins) use $k_r{=}8$; biomarker result-line queries use small $k_r$ (e.g., $4$) since evidence is localized.

\paragraph{2.~LLM Synthesizer}
Given a chunk or chunks of text, the \emph{LLM Synthesizer} (e.g., an LLM such as GPT-4) is prompted to extract a structured representation. The prompt typically includes instructions about the schema to be returned. For instance, we might request that the LLM produce a JSON-like object with \texttt{name}, \texttt{dosage}, and \texttt{start\_date} fields for a \texttt{CancerMedication} entity. Concretely:
\begin{enumerate}
    \item \textbf{Concatenate:} The chunk text plus any instructions (e.g., templates, format specification).
    \item \textbf{Infer:} The LLM extracts attribute values from the chunk’s content.
    \item \textbf{Emit:} A structured representation that respects the required fields and data types.
\end{enumerate}

\paragraph{3.~Collators}
A collator accepts multiple partial extractions and merges or filters them into a final, validated structure. Here, a collator is a deterministic post-processing module (not an LLM unless explicitly stated) that canonicalizes values, enforces type/value-set constraints, deduplicates, and resolves conflicts via simple precedence rules. Typical collator functions include:
\begin{itemize}
    \item \emph{Deduplication:} If multiple extractions have the same \texttt{name} and \texttt{date}, unify them.
    \item \emph{Validation:} Check whether attributes match type constraints or known value sets.
    \item \emph{Conflict Resolution:} If two extractions have contradictory attributes, apply domain logic to keep the most recent or plausible instance.
\end{itemize}

Collation can be formalized as: 
\[
\mathrm{Collate}\bigl( \{R_1,\dots,R_m\} \bigr) = \widehat{R},
\]
where $R_i$ are sets of attribute-value pairs from the LLM Synthesizer, and $\widehat{R}$ is a single consolidated set of entity instances. If the collator depends on other entity data (e.g., a known diagnosis date), it can be passed in as an auxiliary input, ensuring consistent domain constraints across different entity types. Collators can be chained together to produce complex logical strategies. 

\subsubsection{Pipeline Types}

We integrate these components into four canonical pipelines for this specific project, each reflecting a distinct approach to retrieving, synthesizing, and merging entity information. The modular components, however, allows us to create even more complex pipelines which are out of the scope of this work:

\paragraph{(a) \texttt{Single-Step Pipeline}} 
A straightforward, single-pass strategy. The pipeline first retrieves all relevant chunks using either vector or regex methods. The LLM Synthesizer then processes each chunk in one shot to produce the entity attributes, and a collator merges the results. This is well suited for entities that appear in well-defined contexts or are consistently mentioned in the text such as biomarkers.

\paragraph{(b) \texttt{Multi-Step Pipeline}}
A multi-stage approach for more complex or heterogeneous entities. First, we \emph{identify} which variants or subtypes are mentioned (e.g., enumerating possible medications). Next, for each identified subtype, we \emph{extract} more detailed attributes with a targeted LLM prompt. The collator then merges these partial extractions. This approach reduces confusion for the LLM when multiple entity types are possible.

\paragraph{(c) \texttt{Topical Extraction}} 
We split a patient’s documents into coherent \emph{topics} (e.g., \emph{radiology reports}, \emph{molecular testing}, \emph{social history}) and systematically process each topic with a specialized prompt. This is particularly useful when distinct sections of text require different domain knowledge or extraction strategies, but must ultimately be combined into a single patient-level record.

\paragraph{(d) \texttt{Sequential Documents Analysis} }
We treat each document as an individual unit, performing retrieval or chunking within that document, then calling the LLM. The collator merges the partial outputs across all documents, preserving the document lineage. This is valuable when the timing and source of information is critical, such as for pathology or surgical reports that must be chronologically tracked.

\paragraph{Pipeline selection policy}
We choose the pipeline by entity complexity and context:
Single-Step for localized cues with few attributes (e.g., \texttt{Biomarker});
Multi-Step for entities with variants and many attributes (e.g., \texttt{Medication});
Topical when sections require specialized prompts (e.g., radiology vs. molecular testing);
Sequential Documents when provenance and chronology are critical (e.g., \texttt{Staging}, surgery timelines).

\subsubsection{Dependency Management}
When the extraction for an entity (e.g., \texttt{Cancer Medication}) requires context from an upstream entity (e.g., \texttt{Primary Cancer Condition}), the pipeline enforces a dependency order among collators. For instance, if the medication collator depends on a confirmed diagnosis date, the pipeline first finalizes the \texttt{Primary Cancer Condition} collations before generating medication instances. This ensures consistent references (e.g., no medication entry without a corresponding diagnosis period).

In summary, our methods combine modular retrieval (via patterns or embeddings), LLMs (for robust text-to-structure synthesis), and domain-driven collation (for validation, deduplication, and dependency resolution). This general design accommodates a wide array of oncology-specific extractions, from straightforward biomarkers to multi-document treatments and staging workflows. Implementation specifics, including representative prompt templates and retrieval parameters, are detailed in Supplementary Sections \textbf{S-2 }and \textbf{S-3}.

\section{Evaluation}
\label{sec:eval}

In this section, we outline our approach to assessing the performance of \texttt{HARMON-E} in the extraction of oncology variables from clinical notes. Our evaluation targets three core objectives: 
\begin{enumerate}
    \item \textbf{Entity-Level Recall:} Confirming that the pipeline correctly identifies all entities recorded in the ground truth.
    \item \textbf{Precision:} Ensuring the pipeline does not introduce false positives, i.e., entities not present in the ground truth.
    \item \textbf{Attribute-Level Accuracy:} Verifying that all attribute values match the ground truth for each correctly extracted entity.
\end{enumerate}
Together, these objectives confirm that the pipeline not only captures all relevant information but does so accurately at both the entity and attribute levels.

\begin{figure}[t]
    \centering
    \includegraphics[width=0.9\linewidth]{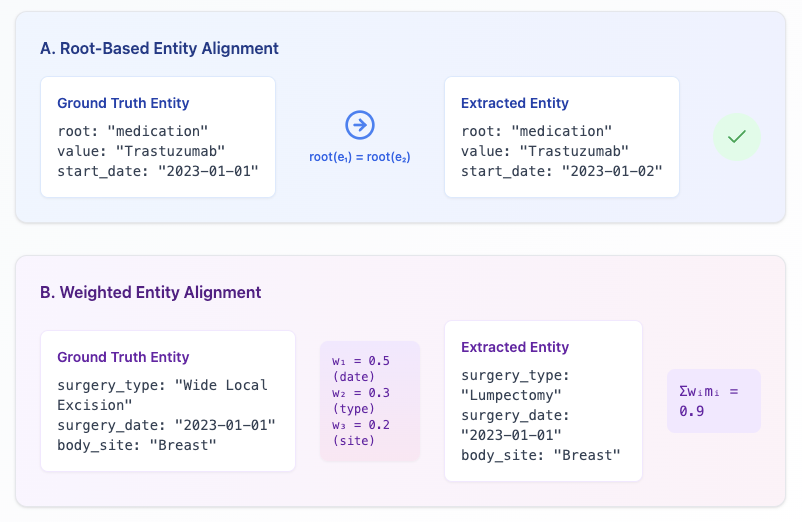}
    \caption{
        \textbf{Overview of Two Entity Alignment Methods.} 
        \textbf{(A)~Root-Based Alignment:} An alignment is established only if both entities share the same \emph{root attribute} 
        (e.g., ``\texttt{medication}'' with value ``Trastuzumab''), making other attributes (such as start dates) irrelevant for basic alignment. 
        \textbf{(B)~Weighted Alignment:} Each attribute (e.g., \texttt{surgery\_type}, \texttt{surgery\_date}, \texttt{body\_site}) 
        contributes a partial score based on a predefined weight. If the sum of matching attributes meets or exceeds a threshold 
        (e.g., 0.9), the ground-truth and predicted entities are aligned.    }    
    \label{fig:entity-alignment}
\end{figure}

\subsection{Entity Alignment Methods}
\label{subsec:alignment-methods}

After running the pipeline, we must determine whether the extracted entities correspond to those in the ground truth data. This alignment step is crucial for measuring entity-level recall, precision, and attribute-level accuracy. Specifically, we employ two methods: Root-based alignment and Weighted alignment.

\subsubsection{Root-Based Entity Alignment}
\label{subsubsec:root-based}
For certain entity types, a specific attribute---the \emph{root attribute}---uniquely identifies that entity. We thus enforce the condition that if these root attributes do not match, the entities cannot be aligned.

\begin{equation}
\text{align}_{\mathrm{root}}(e_1, e_2) \;=\;
\begin{cases}
    1 & \text{if } \mathrm{root}(e_1) = \mathrm{root}(e_2), \\
    0 & \text{otherwise}.
\end{cases}
\label{eq:root-based}
\end{equation}



Root-based alignment applies to entities whose identity is fixed by a single field (Table~\ref{tab:align-schemes}; e.g., \texttt{biomarker\_tested} for Biomarker Summary, \texttt{medication} for Cancer Related Medication).

\subsubsection{Weighted Entity Alignment}
\label{subsubsec:weighted}
For entities where the root attribute may be ambiguous or prone to lexical variation, we adopt a weighted alignment strategy. Each attribute is assigned a weight reflecting its importance in identifying that entity. The alignment score between a ground truth entity $e_1$ and an extracted entity $e_2$ is computed by:

\begin{equation}
\text{align}_{\mathrm{weighted}}(e_1, e_2) \;=\; 
\sum_{i=1}^{k} w_i \cdot \mathrm{match}(a_{1i}, a_{2i}),
\label{eq:weighted-score}
\end{equation}
where $w_i$ is the importance weight of attribute $i$, and $\mathrm{match}(a_{1i}, a_{2i})$ indicates a match of attribute values. A threshold $\tau$ then determines alignment:

\begin{equation}
\mathrm{aligned}(e_1, e_2) \;=\;
\begin{cases}
1 & \text{if } \text{align}_{\mathrm{weighted}}(e_1, e_2) \geq \tau,\\
0 & \text{otherwise}.
\end{cases}
\label{eq:weighted-threshold}
\end{equation}



For entities with lexical variation, we use weighted alignment: attributes receive weights under the guidance of oncology experts, with higher weights on clinically decisive fields (Table~\ref{tab:align-schemes}; e.g., \texttt{surgery\_date}, \texttt{body\_site}) and lower weights on descriptive fields (e.g., \texttt{surgery\_type}). The pairwise score is Eq.~\eqref{eq:weighted-score}; alignment holds when it exceeds a threshold~$\tau$.
\begin{table}[t]
\centering
\footnotesize
\caption{Alignment schemes by entity (anchors and implications).}
\label{tab:align-schemes}
\setlength{\tabcolsep}{4pt}
\renewcommand{\arraystretch}{1.15}
\begin{tabularx}{\textwidth}{
  >{\raggedright\arraybackslash}p{3.0cm}
  >{\centering\arraybackslash}p{1.6cm}
  >{\raggedright\arraybackslash}X
  >{\raggedright\arraybackslash}X}
\toprule
\textbf{Entity} & \textbf{Scheme} & \textbf{Anchor / decisive fields} & \textbf{Implication}\\
\midrule
Biomarker Summary & Root & \texttt{biomarker\_tested} & Different biomarkers never align even if dates match (e.g., \texttt{BRAF} vs \texttt{NRAS}).\\
Cancer Related Medication & Root & \texttt{medication} & Drug names must match; other fields are irrelevant for the root check (e.g., \texttt{Trastuzumab} vs \texttt{Paclitaxel} do not align).\\
Cancer Related Surgery & Weighted & \texttt{surgery\_date}, \texttt{body\_site} \textgreater{} \texttt{surgery\_type} & Decisive fields dominate; lexical variants of type may still align (e.g., “wide local excision” vs “re-excision”).\\
Staging & Weighted & \texttt{stage\_date}, \texttt{stage\_value}, \texttt{stage\_type} & Prioritize date and value; resolves minor notation differences (e.g., \texttt{pT2N0M0} vs \texttt{pT2 N0 M0} align; \texttt{cT2N0M0} vs \texttt{pT2N0M0} do not).\\
\bottomrule
\end{tabularx}
\end{table}


\subsubsection{Entity Alignment Process}
\label{subsubsec:entity-alignment-process}


The entity alignment process follows specific constraints regardless of whether root-based or weighted alignment is employed. The fundamental requirement is the uniqueness constraint, which ensures that each ground truth entity aligns with at most one extracted entity, and vice versa. This one-to-one mapping is essential to prevent artificial inflation of recall metrics.

Once entities are aligned, the process identifies a driver attribute that serves as the anchor for computing entity-level recall and precision metrics. The selection of this driver attribute depends on the alignment method used. In root-based alignment, the root attribute itself—such as \texttt{biomarker\_tested} or \texttt{medication}—naturally serves as the driver. For weighted alignment, the system selects the attribute that received the highest weight assignment as the driver attribute. This driver attribute then becomes the primary reference point for evaluating how well the extraction system captures the essential characteristics of each entity.

\subsection{Manual Evaluation}
\label{subsec:manual-eval}

Beyond automated metrics, we developed a manual evaluation protocol to assess real-world pipeline performance through clinical expert review. This approach acknowledges that divergences between model outputs and ground truth may stem from either genuine model errors or inconsistencies in the original manual abstractions.

Given the resource-intensive nature of comprehensive manual review, we implemented a strategic sampling method that prioritizes cases of disagreement. We quantified divergence through a disagreement scoring mechanism, formally defined as DS(p) in Supplementary Methods S4, that counts mismatched attributes between pipeline outputs and ground truth for each patient. Patients were ranked by their disagreement scores, with the top 50 highest-scoring cases per entity type selected for expert review. This targeted approach concentrates human expertise where it provides maximum insight into system performance limitations.

Clinical abstraction specialists with oncology domain expertise reviewed selected patient records following a standardized protocol. The review process minimized bias by presenting complete patient records without indicators of data source (pipeline vs. ground truth). Reviewers classified each extracted item as correct, incorrect, or missing—categories detailed in Supplementary Methods S4 along with the complete evaluation workflow.

Two complementary performance metrics were derived from this classification: an Acceptance Score and a Missing Rate. These metrics, whose mathematical formulations are provided in Supplementary Methods S4, characterize both extraction accuracy and completeness—critical dimensions for clinical deployment.

\subsection{Dataset}
\noindent \textbf{Dataset Description:} Our study draws upon a real-world dataset of 2{,}250 melanoma patients, each contributing an average of approximately 180 unstructured clinical documents. These documents originated from hundreds of distinct healthcare institutions, reflecting considerable variability in document formatting, language use, and clinical notation styles. In total, the corpus is composed of roughly 50\% HTML-based notes exported directly from the EHR systems, with the remaining 50\% comprising scanned PDF files. Even though the system was tested on Melanoma patients, the same system can be used to get these results by only modifying the valuesets while using same prompts. 

\begin{figure}
    \centering
    \includegraphics[width=0.7\linewidth]{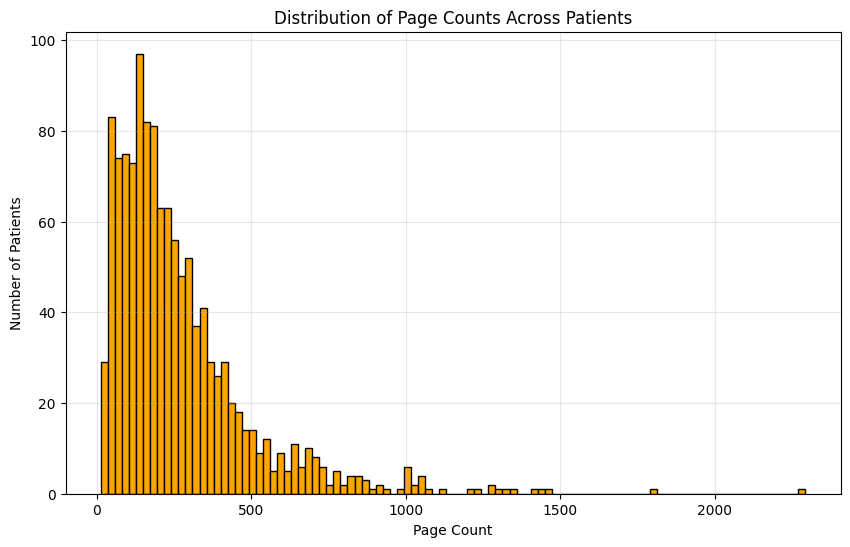}
    \caption{\textbf{Distribution of the Number of Pages per Patient.}
        Each bar represents the number of patients grouped by their total page count, 
        where HTML-based documents are treated as single pages while PDF pages are counted individually. }
    \label{fig:dataset_1}
\end{figure}

To streamline data preparation, all HTML-based documents were treated as single-page entities, irrespective of their actual text length or content density. In contrast, scanned PDF documents were processed at the physical page level, preserving the pagination structure that closely mirrors real-world clinical workflows. In Figure~\ref{fig:dataset_1}, we plot the frequency distribution of per-patient page counts (combining HTML “pages” and PDF pages) for the testing subset. Notably, the distribution exhibits a wide range, with some patients having as few as tens of pages while others have more than a thousand. This spread underscores the variable nature of oncology documentation: some patients receive only a limited number of reports (e.g., routine follow-up or localized treatment), whereas others undergo extensive diagnostic workups, multi-line therapies, and second opinions, producing voluminous records. Such variability in document length and format is precisely what makes the automation of data abstraction both challenging and clinically valuable.

\medskip \noindent \textbf{Ground Truth:} The dataset was curated by a team of qualified oncology data abstractors (ODS-Cs), employing standardized abstraction guidelines refined over a two-year span. The abstractors recorded over one hundred oncology-specific data elements at the patient level, including (but not limited to) primary cancer diagnoses, treatment regimens (e.g., chemotherapies, targeted therapies, immunotherapies), diagnostic test results (e.g., biomarker findings), disease progression events, and comorbidities. This labor-intensive manual curation process allowed for the creation of a rich, high-fidelity “ground truth” dataset, forming the reference standard against which we benchmark our extraction pipeline.

\subsection{Cohort Selection}
\label{subsec:cohort}

We analyzed a retrospective melanoma cohort assembled from participating practices under active data-use agreements. The analytic set comprised \textbf{2{,}250} adult patients with histologically confirmed melanoma and available longitudinal unstructured documentation (clinical notes and scanned reports).

\paragraph{Development–test split.}
The final eligible population (\textbf{$n = 2\,250$} unique patients)
was randomly partitioned into \textbf{50 \%} development
($n = 1\,125$) and \textbf{50 \%} hold‑out test
($n = 1\,125$)\@.  Randomisation was stratified by (i) health‑system
cluster, (ii) year of index diagnosis, and (iii) AJCC stage at
diagnosis to ensure balanced distribution of site‑specific coding
styles and disease severity.

The first half was used iteratively for: \begin{itemize} \item \emph{Prompt Development and Refinement:} Crafting and optimizing large language model (LLM) prompts to ensure comprehensive coverage of target oncology entities and minimize ambiguity in free-text interpretation. \item \emph{Pipeline Configuration and Tuning:} Adjusting individual components—such as retrievers, entity collators, and conflict-resolution logic—to accommodate the diverse formats and terminologies present in both HTML notes and PDF scans. \item \emph{Model Selection:} Comparing the performance of candidate large language models, embedded retrieval methods, and specialized domain rules to choose the pipeline configuration that maximized extraction accuracy while preserving computational efficiency. \end{itemize} Once these elements were established, we used the remaining 1{,}125 patients as our held-out test cohort for final evaluation. This two-phase approach (development and test) was designed to prevent data leakage and ensure that performance metrics accurately reflect the system's ability to generalize to new patients and institutions.

Table S1 compares key baseline characteristics of the hold‑out cohort
against the SEER melanoma registry.


\section{Results}
\label{sec:results}



\begin{figure}
    \centering
    \includegraphics[width=0.74\linewidth]{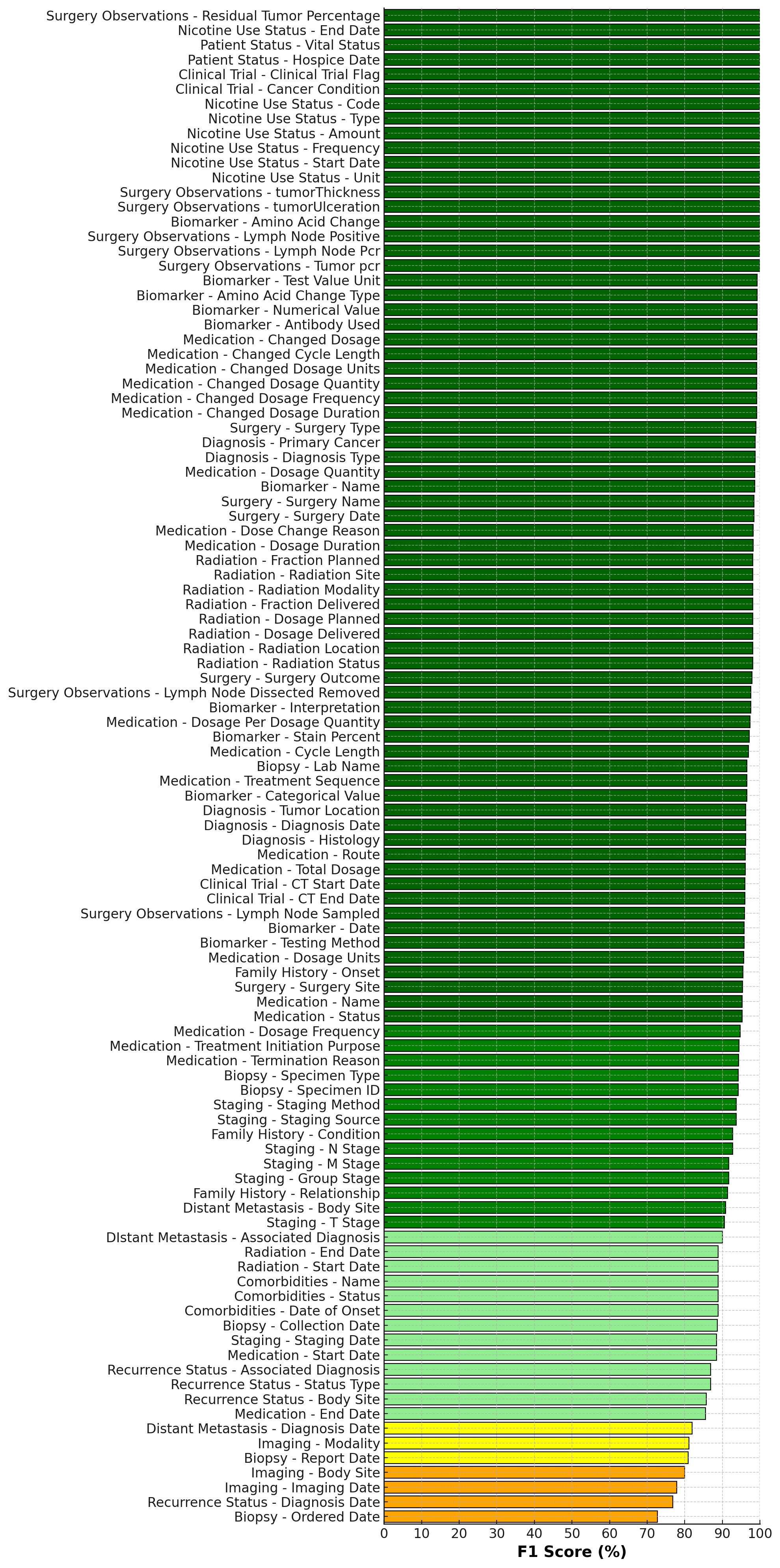}
    \caption{F1-Scores of the evaluated attributes}
    \label{fig:f1scores}
\end{figure}

We summarize in Table~\ref{tab:results} the automated evaluation results for each \emph{Entity} averaged over all its attributes. For each entity, we report \textbf{Precision}, \textbf{Recall}, and \textbf{F1}-score, computed according to the alignment methods described in Section~\ref{subsec:alignment-methods}. Furthermore, we present attribute level results in Fig ~\ref{fig:f1scores}. Overall, the pipeline demonstrates robust performance across a wide range of oncology-related attributes, frequently achieving F1-scores above 90\%. Below, we highlight notable trends and address a few areas with relatively lower scores.

\begin{table}[htbp]
\centering
\caption{Averaged Results For Different Entities}
\label{tab:results}
\scriptsize 
\rowcolors{2}{gray!10}{white} 
\begin{tabular}{p{4cm} p{2.5cm} p{2.5cm} p{2.5cm}}
\toprule
\rowcolor{blue!20} 

         Entity Name &  Precision &  Recall &     F1 \\
\midrule
           Biomarker &     0.9890 &  0.9722 &  0.9806 \\
              Biopsy &     0.8953 &  0.8631 &  0.8789 \\
      Clinical Trial &     1.0000 &  0.9615 &  0.9800 \\
       Comorbidities &     0.8000 &  1.0000 &  0.8889 \\
  Distant Metastasis &     1.0000 &  0.8182 &  0.9000 \\
           Diagnosis &     0.9846 &  0.9600 &  0.9722 \\
      Family History &     1.0000 &  0.8738 &  0.9323 \\
             Imaging &     0.6940 &  0.9337 &  0.7962 \\
          Medication &     0.9722 &  0.9525 &  0.9620 \\
 Nicotine Use Status &     1.0000 &  1.0000 &  1.0000 \\
      Patient Status &     1.0000 &  1.0000 &  1.0000 \\
           Radiation &     0.9481 &  0.9778 &  0.9627 \\
   Recurrence Status &     0.9651 &  0.7445 &  0.8405 \\
             Staging &     0.8771 &  0.9625 &  0.9178 \\
             Surgery &     0.9937 &  0.9633 &  0.9782 \\
Surgery Observations &     0.9939 &  0.9899 &  0.9919 \\
\bottomrule
\end{tabular}
\end{table}

\begin{table}[htbp]
\centering
\caption{Patient‑level performance on the hold‑out cohort
($n=1\,125$).  Values are macro‑averages across the 16 entities;  
$\pm$ denotes 95\% confidence intervals were computed using the Bag-of-Little-Bootstraps (BLB): 10 subsets of size m=128 and 100 bootstrap replicates per subset}
\label{tab:baseline_table_patient}
\scriptsize
\rowcolors{2}{gray!10}{white}
\begin{tabular}{p{4cm} p{2.5cm} p{2.5cm} p{2.5cm}}
\toprule
\rowcolor{blue!20}
\textbf{Configuration} & \textbf{Precision(\%)} & \textbf{Recall(\%)} & \textbf{F1 (\%)} \\
\midrule

GPT‑4o Single‑Step& 78.2$\pm$0.6 & 70.7$\pm$0.7 & 74.3 $\pm$0.6 \\
HARMON-E (w/o collator) & 82.1$\pm$0.5 & 90.7$\pm$0.6 & 86.2$\pm$0.5 \\
\textbf{HARMON‑E (full)} &\textbf{94.1$\pm$0.4} & \textbf{92.4 $\pm$0.5} & \textbf{93.2$\pm$0.4} \\
\bottomrule
\end{tabular}
\end{table}

\subsection{Baselines and Ablation Analysis}
\label{subsec:baselines}

Robust benchmarking of patient‑level extraction is challenging because
most prior work in clinical NLP reports \emph{note‑} or
\emph{sentence‑level} metrics.  Our gold standard dataset enables the \textbf{patient-level} validation, which is closer to real-world use cases.  We
 constructed three reference configurations that satisfy the
same output schema and unit‑of‑analysis:

\begin{enumerate}
  \item \textbf{GPT‑4o Single‑Step.}  
        All notes for a patient are concatenated (truncated to
        32 k tokens) and passed to GPT‑4o
        (June‑2024 snapshot) with a single prompt requesting the full
        JSON schema.  No retrieval, no self‑reflection, no collator.
  \item \textbf{No‑Collator Ablation.}  
        Identical to HARMON‑E except consolidation, validation, and
        dependency‑resolution rules are disabled; LLM generations are
        used “as‑is.”

      \item \textbf{HARMON-E.}  
        Full HARMON-E pipeline with distinct configuration for each entity
\end{enumerate}


All baselines were run on the same 1\,125‑patient hold‑out cohort.
Metrics are macro‑averaged across the 16 entities, 95\% confidence intervals were computed using the Bag-of-Little-Bootstraps (BLB) due to compute constraints: 10 subsets of size m=128 ($\approx N^{0.7}$) and 100 bootstrap replicates per subset with multinomial weights to size N; percentile intervals aggregated across subsets.

Table~\ref{tab:baseline_table_patient} presents precision, recall and
F1.  HARMON‑E exceeds the GPT‑4o single‑step by \(+20\) percentage
points (pp) in F1 (\(p<10^{-3}\)), driven mainly by improvements in date‑sensitive attributes (Medication +28.6 pp). Disabling the collator degrades macro‑F1 by 6.3 pp, confirming the
value of our dependency‑aware post‑processing.  

The GPT‑4o baseline required a \(\sim\)5\(\times\) larger average
prompt (29.4 k vs 6.1 k tokens per patient) and 11.2× higher
inference‑time latency (median = 94 s vs 8.4 s).  Token savings are principally from targeted retrieval.

\subsection{Entity-Level Performance}
\label{subsec:results-entities}

\paragraph{Diagnosis-Related Entities}
For \texttt{Diagnosis} (\texttt{diagnosisType}, \texttt{primaryCancer}, \texttt{diagnosisDate}, \texttt{tumorLocation}, \texttt{histology}), the pipeline achieves F1-scores of 95--98\%. Diagnosis dates are occasionally challenging, but the model still maintains over 95\% F1, a testament to the pipeline’s date normalization and validation steps.

The \texttt{Biomarker} entity, including attributes such as \texttt{name}, \texttt{testingMethod}, \texttt{date}, \texttt{interpretation}, \texttt{categoricalValue}, and \texttt{numericalValue}, consistently shows F1-scores in the mid-to-upper 90\%s. Amino acid change details (\texttt{aminoAcidChange}, \texttt{aminoAcidChangeType}) also approach or exceed 98\%. This underscores the pipeline’s robustness in capturing structured molecular test findings from sometimes lengthy pathology or genetic reports.

TNM staging attributes (\texttt{tStage}, \texttt{nStage}, \texttt{mStage}, \texttt{groupStage}) have F1-scores generally between 90\% and 95\%. \texttt{stagingDate}, however, is relatively lower (88.42\% F1). We observe that temporal references to staging can appear in summary paragraphs or at multiple time points, leading to partial confusion.

\texttt{Recurrence Status} attributes (\texttt{statusType}, \texttt{associatedDiagnosis}, \texttt{diagnosisDate}, \texttt{bodySite}) exhibit moderate-to-high performance, with \texttt{statusType} reaching 86.87\% F1 and \texttt{diagnosisDate} at 76.77\%. \texttt{Distant Metastasis} attributes (e.g., \texttt{associatedDiagnosis}, \texttt{diagnosisDate}, \texttt{bodySite}) consistently outperform recurrence, hovering near 90\%. The slight drop in \texttt{diagnosisDate} for recurrence stems from complexities in distinguishing the actual date of recurrence which can be inferred from imaging, biopsy or even clinician judgement. If the margin of error is increased to +-14 days, the system achieves $\approx 89\%$ F-1 score.

\paragraph{Treatment-Related Entities}

The \texttt{Cancer Related Medication} entity exhibits consistently strong results, with many attributes (\texttt{name}, \texttt{treatmentSequence}, \texttt{route}, \texttt{dosageQuantity}, etc.) achieving F1-scores of 95\% or higher. Several fields (\texttt{changedDosageQuantity}, \texttt{changedDosage}, \texttt{changedDosageUnits}, and related attributes) reach or exceed 99\% F1, indicating that the pipeline accurately captures even nuanced treatment modifications. 

By contrast, date-related attributes (e.g., \texttt{startDate}, \texttt{endDate}) exhibit slightly lower performance. For instance, \texttt{endDate} achieves an F1-score of 85.47\%. While still high, This aligns with known difficulties in extracting date information from unstructured text, potentially due to ambiguous or partial documentation. If the margin of error is increased to +-7 days, the system achieves $\approx 92\%$ highlighting the ability of the system to land at a nearby date, even if not completely correct. This is due to the difficulties in analyzing the date of the last dosage of the medication which is very often derived and not explicitly stated.

The pipeline excels in \texttt{Surgery}, where \texttt{surgeryType}, \texttt{surgeryName}, \texttt{surgeryDate}, and \texttt{surgeryOutcome} all exceed 95\% F1, indicating robust detection of procedural information. \texttt{surgerySite} stands at 95.34\% F1, which still signifies strong performance.

Under \texttt{Surgery Observations}, most attributes (e.g., \texttt{tumorThickness}, \texttt{tumorPCR}, \texttt{lymphNodePCR}) are extracted with near-perfect precision and recall. Even for more complex fields like \texttt{lymphNodeSampled} (\(\sim96\%\) F1), the pipeline demonstrates minimal error rates, suggesting that specialized prompts for pathology details are effective.

\texttt{Radiation} attributes (\texttt{radiationModality}, \texttt{radiationSite}, \texttt{dosageDelivered}, etc.) frequently approach 98--99\% in F1, demonstrating high precision and recall. The date-related fields (\texttt{startDate}, \texttt{endDate}) again pose slight challenges (both at 88.89\% F1), but remain in a range considered acceptable for large-scale automated extractions.

\paragraph{Other Entities}
The attributes \texttt{name}, \texttt{status}, \texttt{dateOfOnset} under the \texttt{Comorbidities} entity exhibit solid performance near 88.89\% F1. Although lower than some other entities, the pipeline still captures the majority of comorbidity data correctly despite variability in how clinical notes reference chronic conditions.

\texttt{Family History} attributes---\texttt{relationship}, \texttt{condition}, and \texttt{onset}---achieve F1-scores in the low-to-mid 90\% range. The pipeline occasionally struggles with ambiguous family relationships or missing explicit onset dates, but it still yields a high level of accuracy overall.

\subsection{Summary of Automated Evaluation}
\label{subsec:results-summary}
In conclusion, the 73 out of 103 \emph{Entity--Attribute} pairs surpass 90\% F1-score, highlighting the effectiveness of the agentic multi-step approach outlined in Section ~\ref{sec:architecture}. Entities with more complex or date-dependent attributes (e.g., certain \texttt{Biopsy} fields, some \texttt{Imaging} details) see moderate performance dips, underscoring the inherent challenges of inconsistent or ambiguous date references in clinical text. Nonetheless, these results illustrate that the proposed \texttt{HARMON-E} pipeline provides robust and comprehensive extraction of oncology-specific data, setting the stage for efficient downstream curation and analysis.

\section{Discussion}
\label{sec:discussion}

Our proposed \texttt{HARMON-E} pipeline systematically addresses the challenge of extracting fine-grained oncology data from heterogeneous EHR notes. The strong performance reported in Table~\ref{tab:results}, with a approximately 75\% \textit{Entity--Attribute} pairs exceeding 90\% F1-score, underscores the effectiveness of a multi-step, agentic approach to clinical information extraction. Here, we examine the factors behind these results, the limitations inherent to current methods, and potential avenues for future refinement.

\subsection{Robustness of the Agentic Approach}
The crux of our pipeline’s success lies in the hierarchical and \emph{modular} decomposition of complex tasks. Rather than forcing a single model to infer all oncology attributes from large, fragmented documents, we divide the workload into discrete steps such as \textit{retrieval}, \textit{LLM-based synthesis}, and \textit{collation}. This strategy closely mimics human workflows where different abstraction tasks (e.g., medications vs.~biomarkers) demand different retrieval contexts and domain-specific prompts.

The high scores across most medication attributes, notably around dosage adjustments and medication status, point to the efficacy of using agentic iterative prompts. In addition, the pipeline’s ability to detect multiple lines of therapy and handle partial or conflicting data suggests that multi-step retrieval—where each pass narrows the focus—reduces confusion that might otherwise arise in single-shot extractions. 

\subsection{Why not ClinicalBERT/SciBERT?}
Those models (and their fine‑tuned variants) emit token‑level BIO tags
or short relation triples per document; transforming such outputs into
patient‑level, longitudinal abstractions requires bespoke heuristics
(e.g., cross‑document clustering, conflict resolution, temporal
alignment) that vary across data partners and are \emph{not} publicly
standardised.  Any comparison would therefore conflate intrinsic model
performance with pipeline engineering choices and is unlikely to offer
actionable insights. Despite that, we ran additional experiments with self-designed postprocessing script and have summarized the results in \textbf{Supplementary S7}.

\begin{figure}[htbp]
    \centering
    \includegraphics[width=\textwidth]{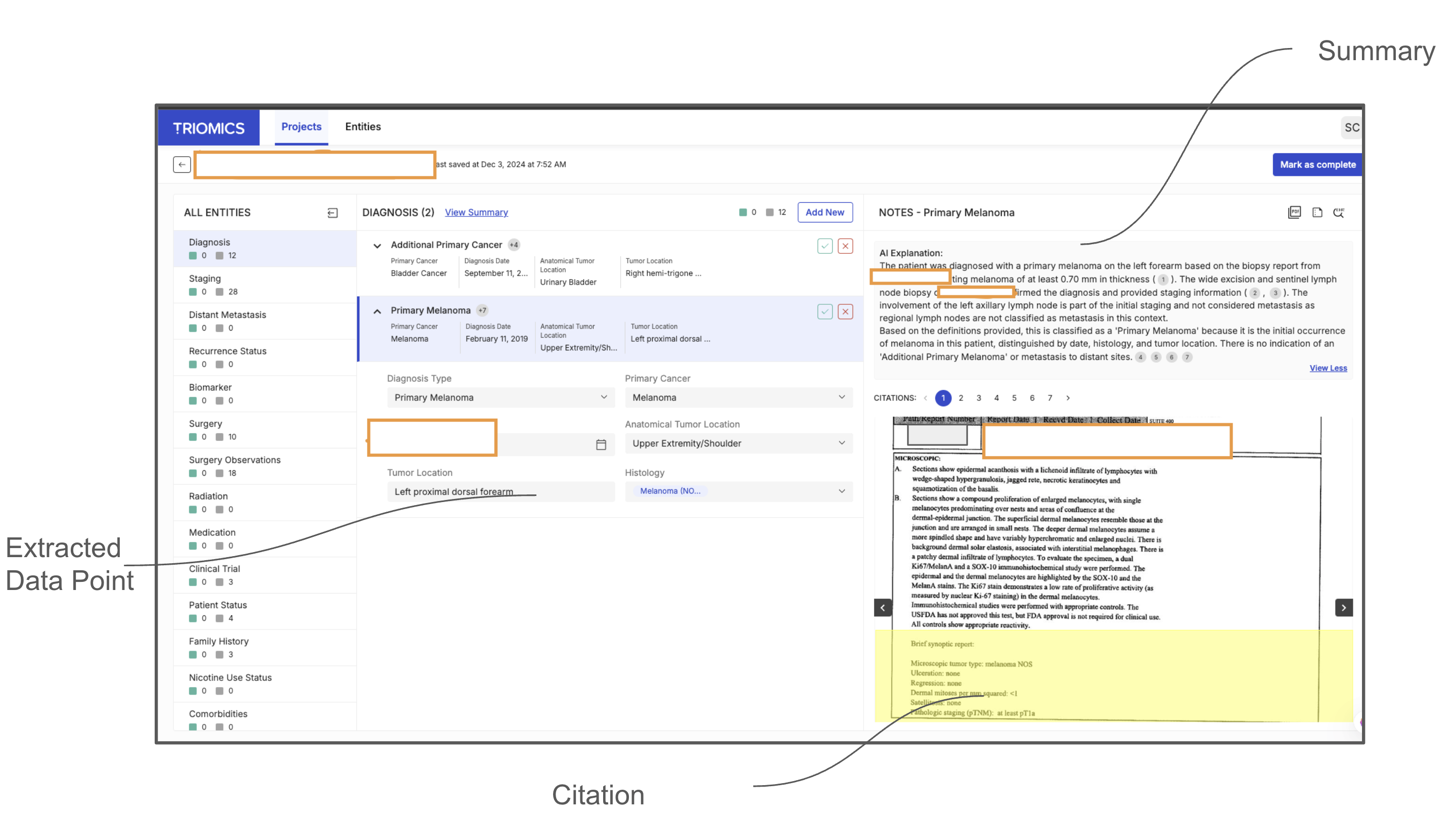}
    \caption{\textbf{Manual adjudication interface for expert validation of HARMON-E extractions.} 
    The data curation platform displays extracted clinical entities in a three-panel layout: 
    (left) hierarchical list of all extracted entities organized by type (Diagnosis, Staging, Distant Metastasis, etc.) with entity counts and completion indicators; 
    (center) detailed view of the selected entity showing all extracted attributes, with the example showing a Primary Melanoma diagnosis dated February 11, 2019 (Deidentified/Shifted date), at anatomical location ``Upper Extremity/Shoulder''; 
    (right) source clinical note with automatic highlighting of the relevant text passage from which the information was extracted, enabling traceable validation. 
    The highlighted yellow section shows the exact source text supporting the extraction, while the orange boxes indicate the specific data points under review. 
    Clinical experts can directly approve extractions using the ``Mark as complete'' button, edit incorrect values inline, or add missing information not captured by the automated pipeline. 
    This interface facilitated the review of 13,609 data points across 50 high-disagreement patients, achieving a 94.1\% direct approval rate as described in Section~\ref{subsec:manual-adjudication}.}
    \label{fig:adjudication-interface}
\end{figure}
\begin{figure}[htbp]
    \centering
    \includegraphics[width=0.8\linewidth]{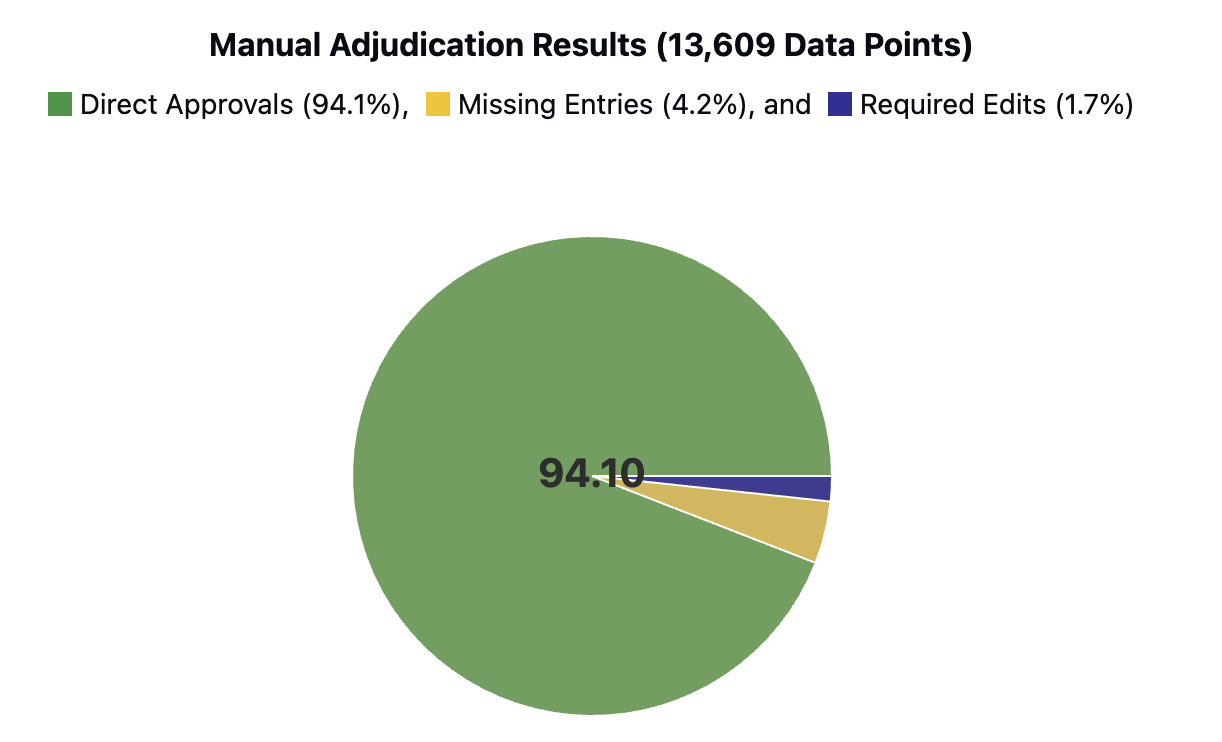}
    \caption{\texttt{HARMON-E} demonstrates high direct-approval rates (94.1\%), with low missing (4.2\%) and required-edit (1.7\%) rates further corrobarting the automated metrics.}
    \label{fig:adjudication-dashboard}
\end{figure}
\subsection{Manual Adjudication Results}
\label{subsec:manual-adjudication}
To complement the automated evaluations, we integrated the pipeline outputs into a data curation platform (as shown in Fig. \ref{fig:adjudication-interface}) and conducted a manual review, or \textit{adjudication}, of 13{,}609 individual data points spanning \texttt{Radiation}, \texttt{Surgery}, \texttt{Medication}, \texttt{Staging}, and \texttt{Diagnosis} entities. As shown in Figure~\ref{fig:adjudication-dashboard} (a representative dashboard excerpt), reviewers with oncology data abstraction expertise assessed each extracted entry in a patient’s record. Specifically, they had three options for each item: 
\begin{itemize}
    \item \textbf{Approve (Correct)}: The pipeline output required no edits.
    \item \textbf{Edit (Incorrect)}: The pipeline output existed but needed to be modified (e.g., a wrong date or attribute value).
    \item \textbf{Add (Missing)}: The pipeline missed an item or field that reviewers deemed relevant.
\end{itemize}

The adjudication process demonstrated that \textbf{94.1\%} of extracted items were \emph{directly approved} without changes. This high approval rate highlights the real-world robustness of the pipeline. In contrast, \textbf{4.2\%} of data points were categorized as \emph{missing entries}, signifying that the pipeline had overlooked relevant details. An additional \textbf{1.7\%} required \emph{edits} to correct partially incorrect information. 

Figure~\ref{fig:adjudication-dashboard} also breaks down the adjudication outcomes by entity type. \texttt{Medication}, \texttt{Radiation}, and \texttt{Diagnosis} exhibit particularly high direct approvals (routinely over 90\%). Where missing data did occur, it was often tied to ambiguous dates or undocumented changes. Edits tended to involve small discrepancies such as inaccurate route or dosage in medication entries or minor staging detail mismatches.

\subsection{Challenges with Date Fields and Context Ambiguity}
Despite the overall strong performance, several date-related attributes (\texttt{startDate}, \texttt{endDate}, \texttt{stagingDate}, etc.) manifest lower recall or precision relative to simpler categorical fields. This is partly because real-world oncology notes often contain multiple, sometimes conflicting date references: for instance, the medication’s original \textit{prescription date} may differ from the \textit{administration start date}, and not all providers consistently document time points in an unambiguous format. 

Similarly, entities like \texttt{Imaging} can appear sporadically across radiology reports, discharge summaries, and referral letters, each containing references to different imaging sessions. The pipeline may over-detect or conflate sessions if a single retrieval chunk contains synonyms or abbreviations pointing to multiple studies. Further refinement of chunking strategies and more sophisticated context tracking could address these limitations, for example by leveraging specialized date resolution modules or advanced temporal reasoning prompts within the LLM.

\subsection{Implications for Clinical Research and Real-World Evidence}
The breadth and depth of high-accuracy extraction demonstrated by \texttt{HARMON-E} show promise for scaling up real-world data (RWD) curation in oncology. Automating the abstraction of TNM staging, biomarker statuses, and multi-line therapies has the potential to expedite clinical trial matching, pharmacovigilance, and precision medicine initiatives. Indeed, if integrated into routine EHR workflows, such an agentic pipeline could accelerate retrospective analyses of large cancer cohorts while maintaining data fidelity akin to manual chart reviews.

Moreover, by capturing a broad range of attributes (e.g., \texttt{familyHistory}, \texttt{comorbidities}, \texttt{nicotineUseStatus}), the system can support comprehensive epidemiological studies that hinge on robust phenotypic representations. The precision seen in medication changes (e.g., \texttt{dosageQuantity}, \texttt{doseChangeReason}) also paves the way for granular analysis of treatment patterns over time.





\section{Conclusion}
\label{sec:conclusion}

We presented \texttt{HARMON-E}, a modular, agentic framework for extracting structured oncology data from heterogeneous, often voluminous EHR notes. By combining domain-aware retrieval, large language models for context-sensitive synthesis, and robust collation, \texttt{HARMON-E} demonstrates state-of-the-art performance on a comprehensive range of oncology related clinical attributes. Automated evaluation highlights F1-scores exceeding 90\% for most attributes, and our multi-step strategy effectively addresses the challenges of conflicting or fragmented documentation.

In automating the generation of curated, patient-level oncology data, \texttt{HARMON-E} offers an impactful tool for both clinical research and operational workflows. Our ongoing efforts focus on extending the pipeline’s applicability to other solid tumors, refining date extraction modules, and incorporating active learning mechanisms for continuous improvement. Ultimately, by reducing the resource-intensive burden of manual abstraction, this framework stands to accelerate real-world evidence generation and support more personalized cancer care at scale.

\section*{Declarations}
\subsection*{Data Availability}
This study is retrospective, and no new data was generated. Due to access restrictions and the risk of re-identification, study data will not be shared externally.

\subsection*{Code Availability}
Pipeline configuration files, pseudo-code for helper scripts and representative prompt templates are released in the Supplementary. The authors agree to provide code snippets at a reasonable request by non-competing entity.

\subsection*{Acknowledgements}
The authors gratefully acknowledge the oncology data abstraction and informatics teams at Ontada for their assistance in data access and clinical validation. The authors also appreciate the engineering and data operations teams at Triomics for their work in implementing the HARMON-E infrastructure, including large-scale data ingestion, retrieval optimization, and validation pipelines. 

\subsection*{Author Contributions}
Hrituraj Singh and Yanshan Wang contributed to the conceptualization of the study. Methodology was developed by Hrituraj Singh, and Shashi kant Gupta. Software and implementation were carried out by Shashikant Gupta, Jerrin John Thomas, and Arijeet Pramanik.. All authors reviewed and approved the final manuscript.

\subsection*{Funding}
This research was conducted with institutional support from Triomics and Ontada. No external funding was specifically dedicated to this study.

\subsection*{Competing Interests}
Hrituraj Singh, Shashikant Gupta, Arijeet Pramanik, Regina Schwind and Jerrin John Thomas are employees of Triomics, Inc. 

Lauren Wiener, Avi Raju, Jeremy Kornbluth, and Zhaohui Su are employees of Ontada LLC.

Yanshan Wang declares no competing interests.

\bibliography{sn-bibliography}


\begin{appendices}
\clearpage
\section*{Supplementary Information}
\addcontentsline{toc}{section}{Supplementary Information}

\subsection*{S-1\quad Environment and Dependencies}
\begin{verbatim}
Language models :   gpt-4o-2024-05-13, Qwen2-7B, o1-preview-2024-09-12
Embeddings      :   text-embedding-3-large
Vector store    :   FAISS v1.8.1  (Inner Product)
Python          :   3.11.4
openai          :   1.26.0
pydantic        :   2.8.1
tiktoken        :   0.6.0
rich            :   13.7.0
faiss-cpu       :   1.8.1
\end{verbatim}

\noindent
\textit{All LLM calls were executed on an Azure OpenAI private endpoint inside a HIPAA-compliant VNet; no PHI left the secure boundary.}

\subsection*{S-2\quad Prompt Templates}

\renewcommand{\arraystretch}{1.2}
\begin{table}[h!]
  \centering\small
  \begin{tabular}{p{2.6cm} p{10.6cm}}
  \toprule
  \textbf{Stage} & \textbf{Representative Template Fragment\footnotemark} \\
  \midrule
  \textbf{Biomarker} \newline \textbf{Single-Step} &
  \verb|system: You are an expert molecular pathologist …|\newline
  \verb|user  : <<SNIPPET>>|\newline
  \verb|assistant: Return JSON {biomarker_tested,…,test_date}.| \\[4pt]

  \textbf{Medication} \newline \textbf{Multi-Step–1} \newline (enumerate drugs) &
  \verb|system: You are an oncology pharmacist …|\newline
   \verb|List all distinct systemic agents mentioned in the snippet.| \\[4pt]

  \textbf{Medication} \newline \textbf{Multi-Step–2} \newline (attributes per drug) &
  \verb|system: For the drug "{{DRUG}}" extract| \verb|{start_date,end_date,route,dose_change_reason}.| \\[4pt]

  \textbf{Collation /} \newline \textbf{Deduplication} &
  \verb|system: Merge entries if medication + start_date| \verb|differ by <=7 days OR share identical event_id.| \\[4pt]

  \textbf{Self-reflection} &
  \verb|assistant: If any required attribute is NULL,| \verb|re-read context and attempt one retry.| \\
  \bottomrule
  \end{tabular}
  \caption*{To ensure methodological transparency while respecting intellectual property considerations, abbreviated versions of the prompts used in this study are provided in the supplementary materials. These abbreviated prompts capture the essential structure and intent of the interactions with the large language model, sufficient for understanding our methodology. Each entity in our synthesis pipeline underwent multiple processing stages, each requiring distinct prompting strategies. Complete prompt templates with full instructions, examples, and specific parameters can be made available to researchers seeking to replicate our findings upon reasonable request and execution of a non-disclosure agreement. Interested parties may contact the corresponding author for access to these materials.}
\end{table}
\footnotetext{Temperature = 0, \texttt{top\_p} = 0.1 for all calls.}

\subsection*{S-3\quad End-to-End Pipeline Configuration}

\lstset{language=json,basicstyle=\ttfamily\footnotesize,frame=single}
\begin{lstlisting}
{
  "pipeline_name": "harmon-e_melanoma_v1",
  "entities": [
    {
      "name": "Biomarker",
      "retriever": {
        "type": "vector",
        "embedding_model": "text-embedding-3-large",
        "k": 12,
        "query_template":
          "Find passages describing laboratory or genomic tests for melanoma."
      },
      "synthesizer": {
        "llm": "gpt-4o-2024-05-13",
        "prompt_file": "prompts/biomarker_single_step.txt",
        "max_tokens": 600
      },
      "collator": {
        "rules": ["deduplicate_by_root: biomarker_tested",
                  "prefer_latest: result_date"]
      }
    },

    {
      "name": "CancerRelatedMedication",
      "retriever": {
        "type": "regex+vector",
        "patterns":
          ["(?i)(nivolumab|pembrolizumab|ipilimumab|vemurafenib)"],
        "k": 20
      },
      "synthesizer": [
        {
          "stage": "enumerate",
          "prompt_file": "prompts/medication_stage1_list.txt"
        },
        {
          "stage": "detail",
          "prompt_file": "prompts/medication_stage2_detail.txt",
          "loop_over": "{{ENUMERATED_DRUGS}}"
        }
      ],
      "collator": {
        "rules": [
          "merge_if_name_and_start<=7d",
          "infer_end_date_from_last_administration",
          "set_status_discontinued_if_end_date<today-28d"
        ]
      }
    }
  ],

  "post_processors": [
    "validate_against_schema",
    "iso8601_date_normalizer",
    "convert_units"
  ],

  "evaluation": {
    "alignment_method": "root_or_weighted",
    "metrics": ["precision", "recall", "f1"],
    "date_tolerance_days": 7
  }
}
\end{lstlisting}

\lstset{language=python}
\begin{lstlisting}
"""
run_harmone.py - Minimal driver to execute the JSON pipeline.
"""
import json, pathlib
from harmone.engine import Pipeline   # lightweight wrapper in SI

cfg_path = pathlib.Path("harmon-e_pipeline.json")
pipe     = Pipeline.from_config(json.loads(cfg_path.read_text()))

for patient_dir in pathlib.Path("/data/melanoma_notes/").iterdir():
    result = pipe.run(
        patient_id = patient_dir.stem,
        note_paths = list(patient_dir.glob("*.txt"))
    )
    pipe.save_json(result, out_dir="outputs/")
\end{lstlisting}

\noindent
This sample configuration file illustrates the structure of a typical HARMON-E pipeline, showing how different modules can be connected and parameterized to create a complete workflow. The accompanying driver script provides a straightforward example of how to load and execute such a pipeline configuration programmatically.

\subsection*{S-4\quad Manual Evaluation Protocol}

This section provides the detailed methodology for the manual evaluation protocol summarized in Section 4.2.

\paragraph{Disagreement-Based Sampling.}
For each patient $p$, we calculated a Disagreement Score $\mathrm{DS}(p)$ representing the total number of mismatched attributes between pipeline outputs and ground truth:
\begin{equation}
\mathrm{DS}(p) = \sum_{j=1}^{N} \delta_j
\end{equation}
where $\delta_j = 1$ if attribute $j$ is mismatched and $0$ otherwise across $N$ total attributes. Patients were ranked by $\mathrm{DS}(p)$ in descending order, with the top 50 highest-scoring patients selected per entity type for expert review.

\paragraph{Review Categories.}
Clinical experts classified each extracted item into one of three categories:
\begin{itemize}
\item \textbf{Correct}: Pipeline output accurately reflects clinical documentation and requires no modification
\item \textbf{Incorrect}: Pipeline output contains errors requiring editing or deletion
\item \textbf{Missing}: Clinically relevant information present in the patient record but absent from pipeline output
\end{itemize}




\begin{table}[h!]
    \centering
    \caption{Step-by-step implementation of the manual evaluation protocol}
    \label{tab:manual-eval-protocol-supp}
    \begin{tabularx}{\linewidth}{@{}>{\centering\arraybackslash}p{0.7cm}
                                        >{\raggedright\arraybackslash}p{3.8cm}
                                        >{\raggedright\arraybackslash}X@{}}
    \toprule
    \textbf{Step} & \textbf{Action} & \textbf{Implementation Details} \\
    \midrule
    1 & Calculate Disagreement Score &
    \begin{itemize}
    \item Compare each attribute across all entity instances
    \item Count mismatches between pipeline and ground truth
    \item Compute $\mathrm{DS}(p)=\sum_{j=1}^{N}\delta_j$
    \end{itemize} \\
    \addlinespace[0.4em]
    2 & Select Review Cohort &
    \begin{itemize}
    \item Sort patients by $\mathrm{DS}(p)$ in descending order
    \item Select top 50 patients per entity type $E\in\mathcal{E}$
    \item Ensure minimum 5 instances per entity type per patient
    \end{itemize} \\
    \addlinespace[0.4em]
    3 & Conduct Blinded Review &
    \begin{itemize}
    \item Present complete patient EHR via curation platform
    \item Remove all indicators of data source (pipeline vs. ground truth)
    \item Expert classifies each item: Correct / Incorrect / Missing
    \item Expert adds any clinically relevant missing information
    \end{itemize} \\
    \addlinespace[0.4em]
    4 & Compute Metrics &
    \begin{itemize}
    \item Tally classifications per patient-entity pair
    \item Calculate Acceptance Score (Eq.~\ref{eq:acceptance-supp})
    \item Calculate Missing Rate (Eq.~\ref{eq:missing-supp})
    \end{itemize} \\
    \bottomrule
    \end{tabularx}
    \end{table}

\paragraph{Performance Metrics.}
The \textbf{Acceptance Score} quantifies the proportion of pipeline-extracted items requiring no modification:
\begin{equation}
\text{Acceptance Score} = \frac{\displaystyle \sum_{p \in P} \sum_{E \in \mathcal{E}} n_{\mathrm{correct}}(p, E)}{\displaystyle \sum_{p \in P} \sum_{E \in \mathcal{E}} n_{\mathrm{extracted}}(p, E)}
\label{eq:acceptance-supp}
\end{equation}

The \textbf{Missing Rate} measures the completeness of extraction:
\begin{equation}
\text{Missing Rate} = \frac{\displaystyle \sum_{p \in P} \sum_{E \in \mathcal{E}} n_{\mathrm{missing}}(p, E)}{\displaystyle \sum_{p \in P} \sum_{E \in \mathcal{E}} \left[n_{\mathrm{extracted}}(p, E) + n_{\mathrm{missing}}(p, E)\right]}
\label{eq:missing-supp}
\end{equation}

\noindent Variable definitions:
\begin{itemize}
\item $P \subseteq \mathcal{P}$: Set of patients selected for review
\item $\mathcal{E}$: Set of entity types evaluated
\item $n_{\mathrm{correct}}(p, E)$: Count of correct extractions for patient $p$, entity type $E$
\item $n_{\mathrm{incorrect}}(p, E)$: Count of incorrect extractions
\item $n_{\mathrm{missing}}(p, E)$: Count of missing items identified by reviewers
\item $n_{\mathrm{extracted}}(p, E) = n_{\mathrm{correct}}(p, E) + n_{\mathrm{incorrect}}(p, E)$: Total extractions
\end{itemize}

\subsection*{S-5\quad Dataset Statistics}
\begin{table}[h!]
\centering\small
\caption{Comparison of baseline characteristics between the SEER 2025 melanoma cohort (2018–2022 diagnoses) and the values reported for the HARMON-E hold-out set.  Differences \,$|\Delta|\!>\!3$ percentage-points are typeset in \textbf{bold}. Some numbers are not exact and have been derived from various sources}
\label{tab:seer_vs_harmone}
\begin{tabular}{@{}lccc@{}}
\toprule
 & \multicolumn{2}{c}{\textbf{Distribution (\%)}} &  \\
\cmidrule(r){2-3}
\textbf{Characteristic} & \textbf{SEER 2025} & \textbf{Dataset} & \textbf{$\Delta$ (pp)}\\
\midrule

\textcolor{blue}{Median age, years\,$^{a}$ }& 66 (IQR 59–74) & \textcolor{blue}{67} & -- \\
Sex – Male & 57.7 & \textcolor{blue}{61.1} & +3.4 \\[2pt]
AJCC Stage I \& II & 77.0 & \textcolor{blue}{48.9} & \textbf{–28.1} \\
AJCC Stage III & 9.5 & \textcolor{blue}{42.5} & \textbf{+33.0} \\
AJCC Stage IV & 4.7 & \textcolor{blue}{1.8} & \textbf{-2.9} \\[2pt]
\bottomrule
\end{tabular}

\vspace{6pt}
\footnotesize
\textit{Notes.} \\
$^{a}$ Age summarised as median (inter-quartile range); no $\Delta$ computed. \\
AJCC = American Joint Committee on Cancer; IQR = inter-quartile range; pp = percentage-points.\\
\end{table}

We extracted reference statistics from SEER public database along with few research papers that summarize melanoma specific statistics. The table \ref{tab:seer_vs_harmone} summarizes the statistics.
\subsection*{S‑6\quad Comparison with Prior Baseline Systems}
\begin{table}[h!]
\centering\small
\caption{Macro‑averaged performance of prior baselines on the hold‑out melanoma cohort.  
Not all pipelines could be modified to approximate all data points in the schema}
\label{tab:baseline}
\rowcolors{2}{gray!10}{white}
\begin{tabular}{p{5.1cm}cccc}
\toprule
\rowcolor{blue!20}
\textbf{Pipeline} & \textbf{Prec.\ (\%)} & \textbf{Rec.\ (\%)} & \textbf{F1\ (\%)} & \textbf{Date Err.$^{*}$ (\%)} \\  
\midrule
cTAKES 4.0 + rule post-proc      & 63.7 & 54.2 & 58.6 & 27.8 \\
SciSpacy + CRF aggregation       & 67.3 & 61.8 & 64.4 & 30.3 \\
ClinicalBERT (fine-tuned, note-level) & 66.5 & 68.9 & 67.7 & 26.2 \\
\addlinespace
\bottomrule
\end{tabular}

\vspace{4pt}
\footnotesize
\emph{Notes.}  
Prec.\,= macro‑precision; Rec.\,= macro‑recall.  
(*) Date Err.\,= percentage of extracted date attributes deviating by \(>\!\pm14\) d from the reference.
\end{table}
To contextualise \texttt{HARMON‑E}’s performance, we re‑implemented three representative extraction pipelines that are commonly cited in clinical‑NLP literature. Not all variables were supported - so we just averaged over whatever could be processed using each method.  
All baselines were executed on the identical 1 125‑patient hold‑out cohort described in Section \ref{subsec:cohort}.  
Because none of the legacy systems natively consolidate information across hundreds of notes, we added a lightweight post‑processor that (i) clusters identical concepts within a 14‑day window and (ii) propagates the most frequent attribute value.  
The configurations and macro‑level results are summarised in Table \ref{tab:baseline}.

\bigskip

\end{appendices}


\clearpage

\end{document}